\documentclass[10pt,twocolumn,letterpaper]{article}

\usepackage[pagenumbers]{cvpr} 

\usepackage{graphicx}
\usepackage{amsmath}
\usepackage{amssymb}
\usepackage{booktabs}

\usepackage{url}            
\usepackage{amsfonts}       
\usepackage{nicefrac}       
\usepackage{microtype}      
\usepackage{enumitem}
\usepackage{amsmath,amssymb,amsbsy}
\usepackage{multirow}
\usepackage{multicol}
\usepackage{wrapfig}
\usepackage{algorithm}
\usepackage{algpseudocode}
\usepackage{caption}
\usepackage{subcaption}

\usepackage[pagebackref,breaklinks,colorlinks]{hyperref}
\usepackage[capitalize]{cleveref}
\crefname{section}{Sec.}{Secs.}
\Crefname{section}{Section}{Sections}
\Crefname{table}{Table}{Tables}
\crefname{table}{Tab.}{Tabs.}

\begin{document}

\title{A biological vision inspired framework for machine perception of abutting grating illusory contours}

\author{Xiao Zhang$^{1}$, Kai-Fu Yang$^{1*}$, Xian-Shi Zhang$^{1}$, Hong-Zhi You$^{1}$ \\ 
Hong-Mei Yan$^{1}$, Yong-Jie Li$^{1}$\thanks{Corresponding author. \\ A peer review version has been accepted by Pattern Recognition. DOI: https://doi.org/10.1016/j.patcog.2026.114185} \\
	\normalsize$^1$ Sichuan Cancer Hospital \& Institute, School of Life Science and Technology, \\ \normalsize University of Electronic Science and Technology of China.
	\\
	\small\texttt{xxyh0201@gmail.com; \{yangkf, liyj\}@uestc.edu.cn}
}

\maketitle

\begin{abstract}
    Higher levels of machine intelligence demand alignment with human perception and cognition. Deep neural networks (DNN) dominated machine intelligence have demonstrated exceptional performance across various real-world tasks. Nevertheless, recent evidence suggests that DNNs fail to perceive illusory contours like the abutting grating, a discrepancy that misaligns with human perception patterns. Departing from previous works, we propose a novel deep network called illusory contour perception network (ICPNet) inspired by the circuits of the visual cortex. In ICPNet, a multi-scale feature projection (MFP) module is designed to extract multi-scale representations. To boost the interaction between feedforward and feedback features, a feature interaction attention module (FIAM) is introduced. Moreover, drawing inspiration from the shape bias observed in human perception, an edge detection task conducted via the edge fusion module (EFM) injects shape constraints that guide the network to concentrate on the foreground. We assess our method on the existing AG-MNIST test set and the AG-Fashion-MNIST test sets constructed by this work. Comprehensive experimental results reveal that ICPNet is significantly more sensitive to abutting grating illusory contours than state-of-the-art models, with notable improvements in top-1 accuracy across various subsets. This work is expected to make a step towards human-level intelligence for DNN-based models.
\end{abstract}

\section{Introduction}
    Contour perception is one of the most important components of visual perception of both human and machine, since contours convey essential information about object shapes and serve as a pivotal cue for figure-ground segregation \cite{RN884}. Typically, contour perception arises at regions where pixel values change abruptly; however, even without stimulation from color contrast or luminance gradients, human visual system can still perceive clear boundaries \cite{RN760}. Such boundaries are defined as illusory contours \cite{RN761}, also known as subjective or anomalous contours \cite{RN775}. As a canonical visual illusion, illusory contours offer an invaluable window into understanding the neural mechanisms and computations underlying visual perception \cite{RN758,RN879,RN882,RN881,RN767,RN880}. Research in cognitive psychology and neurophysiology indicates that the perception of illusory contours involves multiple mechanisms including cortical circuits \cite{RN767,RN792}, visual attention \cite{RN811}, and perceptual organization \cite{RN761, RN774}, constituting an intricate perceptual process. Beyond humans, this perceptual ability is also widely discovered in other biological visual systems, such as rodents \cite{RN792}, non-human primates \cite{RN762}, fish \cite{RN765,RN763}, and birds \cite{RN766}. In a way, the ability of perceiving illusory contours promotes the robustness of visual systems in processing incomplete and ambiguous visual inputs \cite{RN758,RN759,RN932}.

    On the other hand, Deep Neural Networks (DNNs) have achieved and even surpassed human benchmarks in various computer vision tasks, such as image classification \cite{RN745,RN608,RN744} and edge detection \cite{RN734,RN746}. To some extent, the development of DNNs has benefited from the advancements in neuroscience. A clear instance is the hierarchical architecture of DNNs, which is directly inspired by the visual neural mechanisms of cats \cite{RN619, RN344}. Nevertheless, Fan et al. \cite{RN756} and Baker et al. \cite{RN768} have pointed out that current convolutional neural network (CNN) and Transformer-based DNNs lack the ability to perceive illusory contours. For example, in certain cognitive psychology tasks, such as the classification of abutting grating illusion images \cite{RN756}, DNNs still perform poorly. These models fail to account for the experimental results from cognitive psychology \cite{RN754}. Even the state-of-the-art large vision-language models (LVLMs) also show substantial discrepancies with humans in the perception of visual illusions \cite{RN804}.

    Taken together, illusory contour perception is a fundamental capability in biological visual perception. To achieve human-like intelligence, DNNs should also possess such perceptual ability. However, the truth is that a significant misalignment persists between DNNs and humans in terms of perceiving illusory contours, which contradicts the ultimate goal of artificial intelligence (AI) \cite{RN885}. Constructing DNNs models capable of perceiving illusory contours is expected to provide novel insights into utilizing physiological mechanisms to guide the design of neural networks with more human-like intelligence.

    \begin{figure}[t]
        \centering
        \includegraphics[width=80mm]{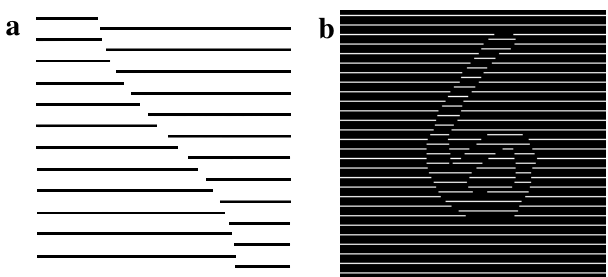}
        \caption{Demonstration of illusory contours. \textbf{a.} An abutting grating illusion with a simple curve, modified from Kanizsa \cite{RN775}. \textbf{b.} An abutting grating illusion with a handwritten digit "6" from the AG-MNIST dataset \cite{RN756}.}
        \label{fig1}
    \end{figure}

    Here, we investigate how to construct a deep neural network with the capability of perceiving abutting grating illusory contours by drawing inspiration from the visual neural mechanisms. Specifically, inspired by inter- and intra-cortical circuits in the visual cortex, we propose a novel deep neural network with feedback connections, termed illusory contour perception network (ICPNet), which demonstrates enhanced perception along the abutting grating illusory contours (see Fig. \ref{fig1} for examples). To the best of our knowledge, ICPNet is the first neural network with a significant ability to perceive the abutting grating illusory contours. Furthermore, rather than relying on a single convolutional layer in the initial layers of networks, a special block is designed to extract multi-scale representations of the input. To effectively integrate features from feedforward and feedback pathways, we then develop a feature interaction attention module (FIAM). Moreover, building upon the shape bias in human visual perception \cite{RN887}, we combine the edge detection task and the image classification task by introducing shape constraints into our network. An intriguing consequence of this multi-task learning framework is that ICPNet exhibits increased focus on global contours and foreground information, aligning with the principles of visual perception \cite{RN887,RN936,RN918}. Ultimately, to enhance the diversity of abutting grating illusory images, we contribute the AG-Fashion-MNIST test set. Comprehensive experimental results on the AG-MNIST and AG-Fashion-MNIST test sets demonstrate that our proposed ICPNet possesses superior illusory contour perception performance compared to other state-of-the-art computational models. 

\section{Related Work}
    \noindent\textbf{Neural mechanisms of illusory contour perception}.
        The neural mechanisms of illusory contour perception have attracted widespread interest in neuroscience. Initially, some scholars argued that such perception relied solely on interaction in early visual cortices, namely the primary visual cortex (V1) and the secondary visual cortex (V2) \cite{RN787,RN888,RN789,RN786,RN783,RN929}. However, in illusory contour figures such as the Kanizsa triangle \cite{RN764}, when the distance between the inducers exceeds a certain threshold, the neurons in V1/V2 no longer respond to the illusory contours \cite{RN783}. In contrast, human vision generally encompasses long-range illusory contour perception \cite{RN797}. Meanwhile, the perception of illusory contours is critically and permanently impaired by higher-order visual cortex lesions \cite{RN790,RN931}. Therefore, it was proposed that illusory contour perception also depends on neurons with larger receptive fields (RF), indicating that higher-order cortical areas mediate the formation of illusory contours \cite{RN930}. Given these reflections, some researchers have elucidated that illusory contour perception is primarily shaped through feedback modulation from higher-order visual cortices, including V4, lateral occipital complex (LOC), and inferior temporal (IT) cortex \cite{RN795,RN792,RN767,RN791,RN796,RN793,RN935}. In contrast to the above-mentioned studies, Cheng and colleagues leveraged DNNs to reconstruct the internal representations underlying illusory perception from functional magnetic resonance imaging (fMRI) data. Their findings indicate that abutting grating illusory contour perception involves a distributed network spanning V1 to V4 and higher-order cortical regions, such as LOC, with the perception mainly localized in early visual cortices \cite{RN773}.
        
        In summary, physiological evidence highlights that illusory contour perception is a constructive process reliant on the interaction between lower- and higher-order visual cortical areas. Feedback modulation plays a crucial role in this perception besides feedforward connections.

    \vspace{0.1cm}
    \noindent\textbf{Illusory contour perception of DNNs}

        \begin{figure*}[t]
            \centering
            \includegraphics[width=110mm]{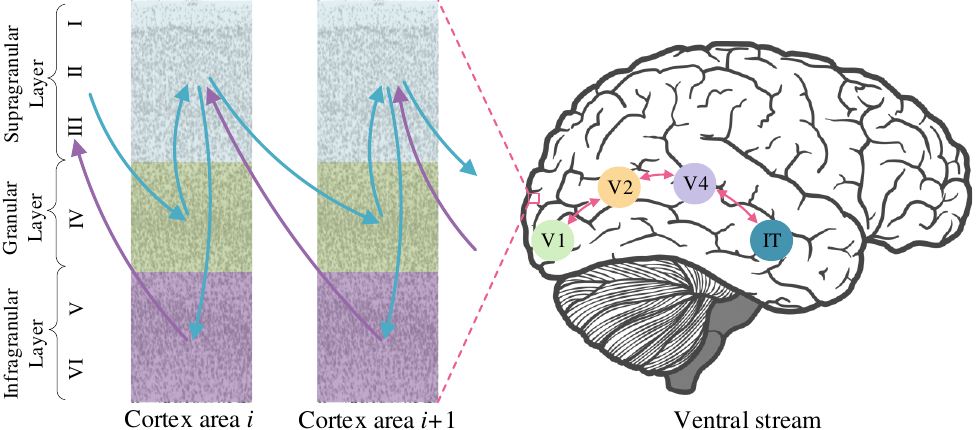}
            \caption{Schematic diagram of ventral stream, cortical structures, and circuits. Blue arrows denote feedforward circuits, while purple arrows represent feedback circuits.
            \label{fig2}}
        \end{figure*}

        Illusory contour perception fulfills its crucial role in enabling biological visual systems to interpret complex scene information \cite{RN761,RN776}. Baker and colleagues \cite{RN768} explored whether DNN models used the same features as humans to perceive real and illusory contours, and concluded that DNNs lacked genuine illusory contour perception. Given that feedforward-based DNNs lacked feedback modulation, Lotter et al. \cite{RN801} observed that the predictive coding \cite{RN806} based on PredNet \cite{RN805} model exhibited dynamic characteristics similar to those of superficial neurons in macaque V1/V2 while processing the Kanizsa square \cite{RN775}. Therefore, they suggested that the predictive coding architecture reveals activation along the illusory contours. Furthermore, Pang et al. \cite{RN803} combined predictive coding with CNN to construct a model capable of perceiving the Kanizsa square. They posited that the illusory contour perception of DNNs arose from self-supervised reconstruction training and feedback connections. Nevertheless, a major drawback of their model is that the classification result is obtained by averaging the outputs of nine sub-models.

        In the aforementioned studies, the training and test images were typically generated by hand-crafted illusory contours popularly presented in psychological literature, such as Kanizsa square \cite{RN775}. These manually designed datasets are limited in quantity, which hinders comprehensive evaluation of illusory contour perception of DNNs. Recently, Fan et al. \cite{RN756} proposed a novel method for converting the MNIST dataset \cite{RN808} into an abutting grating illusory contours dataset, thereby constructing the AG-MNIST test set. Models can be trained on the original MNIST training set and then their illusory contour perception capability can be directly evaluated on the AG-MNIST test set. They found that the performance of the vast majority of DNNs was close to random guessing by evaluating 109 models based on CNN and Transformer architectures, which supports the results of Cheng \cite{RN773}. Differing from all previous works, Zhang et al. \cite{RN804} assessed four LVLMs from the perspective of human-like visual illusion perception. They also observed that these models generally aligned poorly with human visual illusion perception.
        
        Together, existing approaches for modeling illusory contour perception are difficult to transfer to real-world tasks and most current neural networks essentially lack the capability of perceiving illusory contours like the abutting grating. This work aims to construct DNNs with the power to detect illusory contours and distinguish figure from background by learning visual mechanisms, which is expected to help make a step towards human-level intelligence for DNN-based models.

\section{Methodology}
    \subsection{Preliminary}
    \noindent\textbf{Cortical structures and circuits.} Physiological studies have demonstrated that neurons in the primate cortex are organized into a columnar structure comprising six distinct layers (I$\thicksim$VI) \cite{RN893,RN894}, as shown in Fig. \ref{fig2}. Layers I$\thicksim$III, layer IV, and layers V$\thicksim$VI are respectively termed the supragranular layer (SL), granular layer (GL), and infragranular layer (IL). The cortex supports high-order cognitive functions \cite{RN896} and is one of the most intricate structures in the brain. The GL primarily receives inputs from the SL of the previous cortical area and projects information to the SL of the current area \cite{RN897}. Subsequently, the SL transmits information to the pyramidal neurons in the deeper IL \cite{RN896} and receives feedback inputs from higher cortical areas \cite{RN935}.

    Moreover, within the ventral stream (i.e., V1$\rightarrow$V2$\rightarrow$V4$\rightarrow$IT) \cite{RN933}, as the cortical hierarchy progresses, the SL becomes significantly thicker and contains larger pyramidal neurons, accompanied by a thinning of the GL \cite{RN899}. Such change supports the pivotal role of the SL in cortical information integration \cite{RN896}. Although the IL is thinner than the SL, it likewise exhibits this trend \cite{RN934,RN896,RN899}. Overall, this fundamental cortical organization provides the structural basis of our model for illusory contour perception.

\subsection{ICPNet architecture}
    \begin{figure*}[t]
        \centering
        \includegraphics[width=170mm]{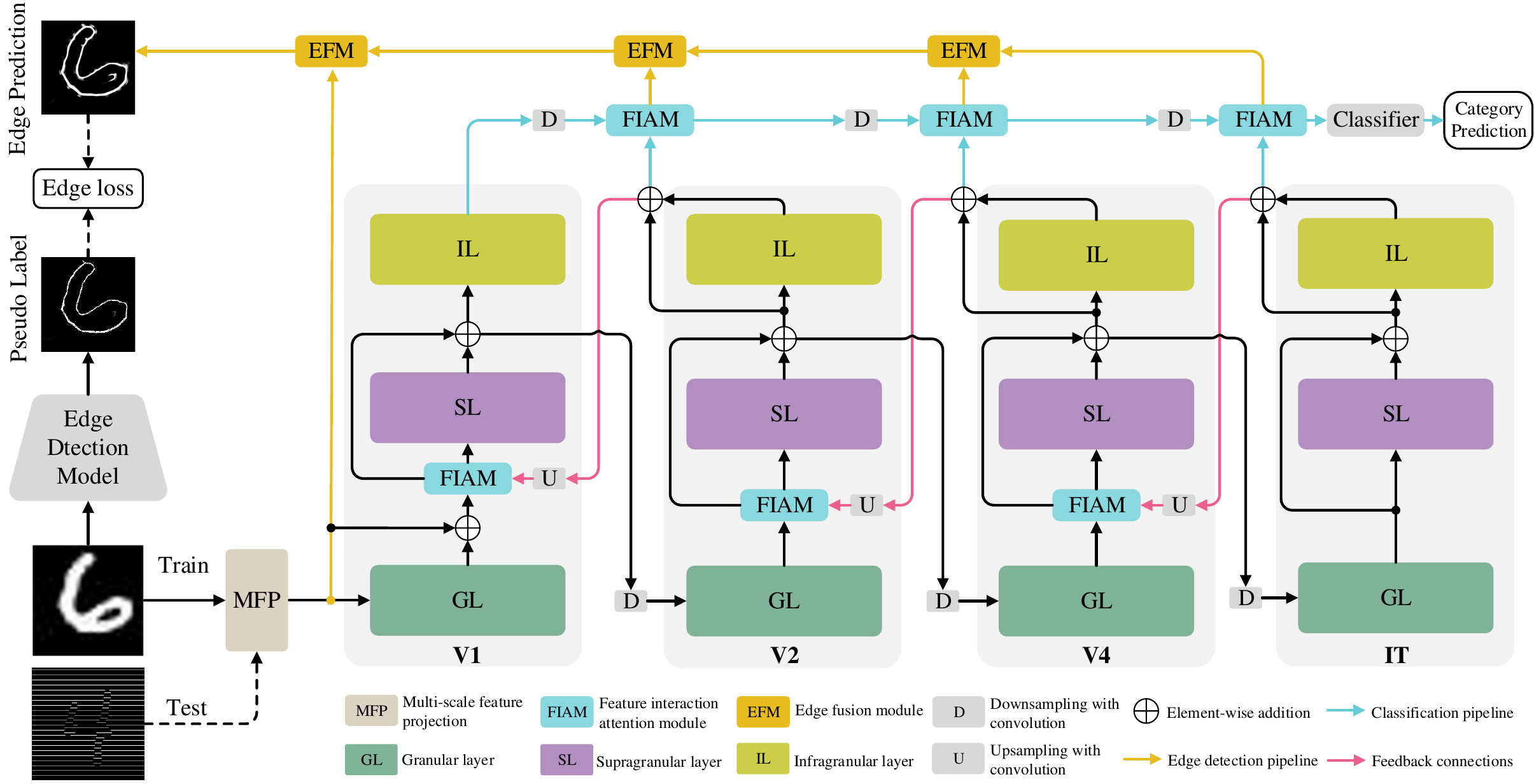}
        \caption{The overall architecture of ICPNet. Our network consists of two main pipelines: classification and edge detection. The former obtains outputs through four stages with feedback connections, while the latter employs a top-down structure to integrate side outputs via EFM to generate edge predictions. During the training phase, we freeze the network weights of the edge detection model pre-trained on the BSDS-VOC dataset \cite{RN627,RN502}.}
        \label{fig3}
    \end{figure*}

    \begin{table*}[t]
        \centering
        \caption{Detailed parameters of four stages with a structure of [kernel size, channel number, dilation rate]. "r", "DW", and "Conv" indicate dilation rate, depthwise separable convolution, and vanilla convolution layer, respectively. "GL", "SL", and "IL" correspond to the granular layer, supragranular layer, and infragranular layer, respectively.}
        \setlength{\tabcolsep}{3pt}
        \renewcommand{\arraystretch}{1.2}
        \begin{tabular}{ccccc}
        \hline
        Block name          & Stage 1                         & Stage 2                          & Stage 3                          & Stage 4                          \\ \hline
        \multirow{3}{*}{SL} & [4$\times$4DW, 128, r=2]                 & [7$\times$7DW, 128, r=2]                  & [9$\times$9DW, 256, r=3]                  & [9$\times$9DW, 512, r=3]                  \\
                            & \multirow{2}{*}{[1$\times$1Conv, 64, r=1]}    & [3$\times$3DW, 256, r=6]                  & [3$\times$3DW, 512, r=12]                 & [3$\times$3DW, 1024, r=12]                \\
                            &                                 & [1$\times$1Conv, 128, r=1]                     & [1$\times$1Conv, 256, r=1]                     & [1$\times$1Conv, 512, r=1]                     \\ \hline
        \multirow{3}{*}{GL} & [5$\times$5DW, 64, r=1]                       & \multirow{3}{*}{[6$\times$6DW, 128, r=2]} & \multirow{3}{*}{[7$\times$7DW, 256, r=3]} & \multirow{3}{*}{[7$\times$7DW, 512, r=3]} \\
                            & [3$\times$3DW, 128, r=2]                 &                                  &                                  &                                  \\
                            & [1$\times$1Conv, 64, r=1]                     &                                  &                                  &                                  \\ \hline
        \multirow{2}{*}{IL} & \multirow{2}{*}{[4$\times$4DW, 64, r=2]} & [7$\times$7DW, 256, r=2]                  & [9$\times$9DW, 512, r=3]                  & [9$\times$9DW, 1024, r=3]                 \\
                            &                                 & [1$\times$1Conv, 128, r=1]                     & [1$\times$1Conv, 256, r=1]                     & [1$\times$1Conv, 512, r=1]                     \\ \hline
        \end{tabular}
        \label{table1}
    \end{table*}

    The overall architecture of ICPNet is illustrated in Fig. \ref{fig3}. This model adopts a multi-task learning framework, which includes classification task and edge detection task. For classification, the input image is first processed by the multi-scale feature projection (MFP) block to obtain multi-scale features. This is followed by four stages inspired by cortical structures and circuits, corresponding to V1, V2, V4, and IT areas, respectively. Each stage consists of three components, including the granular layer (GL), supragranular layer (SL), and infragranular layer (IL). Regarding cortical connectivity, there is a massive network of functional connections both between and within cortical regions. To simplify neural network design, we only consider feedforward and feedback connections between adjacent cortices. Therefore, in addition to feedforward connections, feedback connections with global features are also established to simulate the modulation from V2 to V1 \cite{RN900}, V4 to V2 \cite{RN647}, and IT to V4 \cite{RN919}. The feature interaction attention module (FIAM) is designed to integrate the feedback and feedforward features. Moreover, inspired by the shape bias in the visual system \cite{RN887}, we introduce the edge detection task to inject shape constraints into the network. The edge fusion module (EFM) merges side outputs with different resolutions in a top-down fashion to obtain the edge prediction. By jointly optimizing the both tasks, the network can concentrate more effectively on foreground information.

    With the four stages, the convolutional kernels in each stage are designed according to the RF size of different cortical areas. Table \ref{table1} lists the detailed parameters of the four stages. Concretely, the average RF size of V2 is approximately twice that of V1 \cite{RN901}, and the average RF size of V4 is about twice that of V2 \cite{RN902}. Therefore, in conjunction with the structural rationality of the network, the kernel sizes between stages maintain an approximate double relationship, except for the last two stages. Meanwhile, to control the number of parameters, dilated convolution \cite{RN904} and depthwise separable convolution (DW) \cite{RN863} are utilized in the design of the basic convolutional blocks. Following ResNet architecture \cite{RN608}, the number of channel of four stages are set to 64, 128, 256, and 512, respectively. To align the size and channel number of feature maps, we employ downsampling and upsampling with convolution used in Swin \cite{RN809}. Specifically, the upsampling with convolution is formulated as
    \begin{equation}
        f_u=Conv\left( LN\left( Up\left( \cdot \right) \right) \right),
    \end{equation}
    where $f_u$ is the aligned features. $LN$, $Conv$, and $Up$ denote the layer normalization \cite{RN911}, a 1$\times$1 convolution layer with stride 1, and the transposed convolution, respectively. Replacing bilinear upsampling with pooling results in the downsampling with convolution. Neurophysiological findings reveal that, besides the IT area, subcortical regions also interact with the IL of cortices to facilitate efficient object recognition \cite{RN898,RN748,RN906}. Accordingly, we fuse side outputs from the IL via the FIAM in a bottom-up fashion. The final output is then fed into the classifier to implement classification. The classifier is defined as
    \begin{equation}
        P_{cls}=softmax \left( FC\left( LN\left( GAP\left( \cdot \right) \right) \right) \right),
    \end{equation}
    where $P_{cls}$ is the classification probability. $FC$ and $GAP$ are a fully-connected layer and a global average pooling layer, respectively.

    \subsection{Multi-scale feature projection (MFP)}
        \begin{figure*}[t]
            \centering
            \includegraphics[width=150mm]{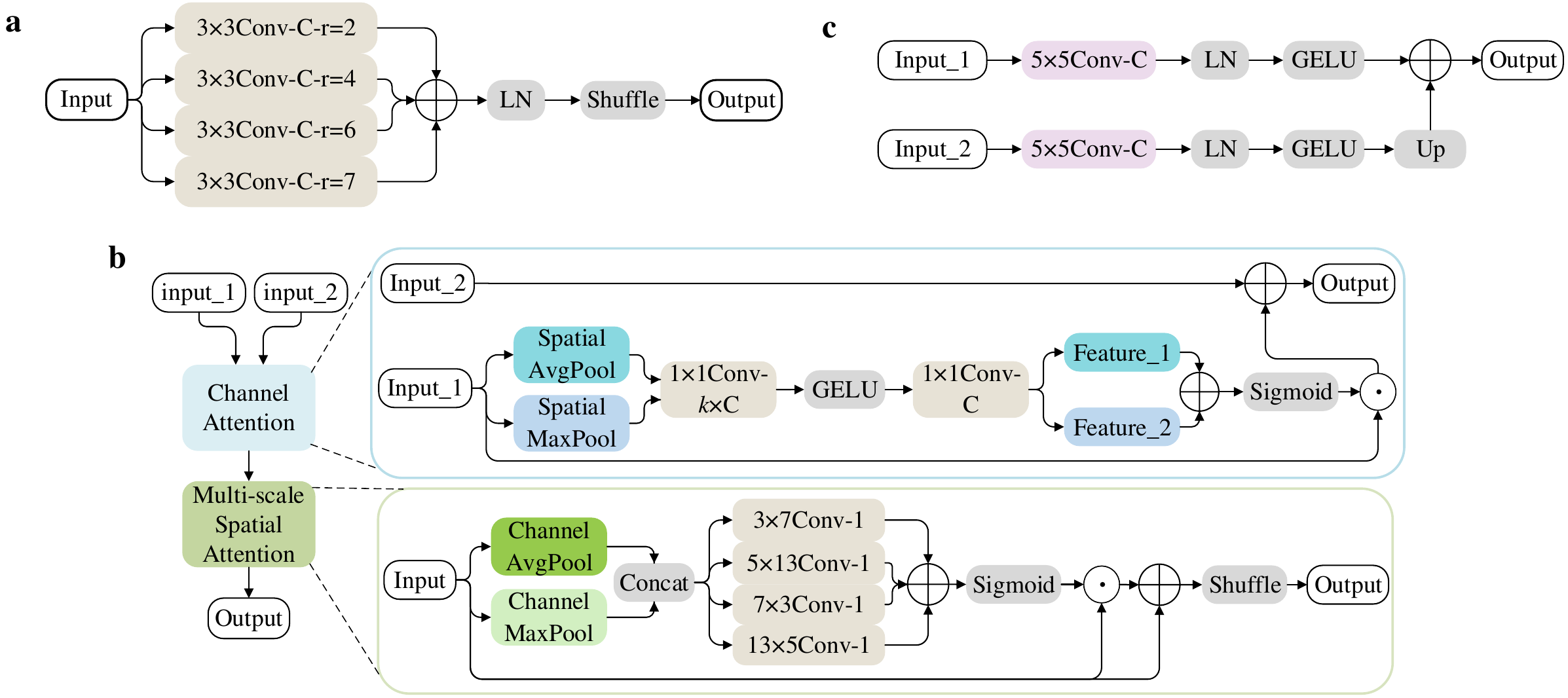}
            \caption{The detailed structures of submodules. \textbf{a.} MFP. "C" is equal to 64. "r" and "$\oplus$" denote dilation rate and element-wise addition, respectively. \textbf{b.} FIAM. FIAM is composed of the channel attention and the multi-scale spatial attention. "input\_1" and "input\_2" indicate the features from the lower-level stage and the higher-level stage, respectively. "$\odot$", "$\mathit{k}$", and "C" denote element-wise multiplication, expansion factor, and channel number, respectively. \textbf{c.} EFM. "Up" denotes the bilinear upsampling.}
            \label{fig4}
        \end{figure*}

        In general, the initial layers of the model focus on how to process images at the beginning of a neural network and typically apply a single-layer convolution to extract initial features. For instance, CovNeXt \cite{RN907} employs a 4$\times$4 non-overlapping convolutional layer with stride 4 as the stem. However, multi-scale features are critical for CNN perception \cite{RN908}. Therefore, we develop MFP with dilated convolution to extract multi-scale representations, as shown in Fig. \ref{fig4}\textcolor{red}{a}. In the MFP block, the input image first passes through convolutional layers with stride 2 and different dilation rates in parallel, and the resulting feature maps are summed together, followed by LN. To further enhance the interaction among the representations, we apply a channel shuffle \cite{RN909} to obtain the final output. Although the MFP module is relatively simple, our ablation studies demonstrate that the multi-scale features is essential for abutting grating illusory contour perception.

    \subsection{Feature interaction attention module (FIAM)}
        As shown in Fig. \ref{fig4}\textcolor{red}{b}, to better integrate features from different levels, we construct the FIAM built upon the CBAM \cite{RN675}. For feedback connections, the features from the higher-level stages (i.e., input\_2) first undergo upsampling with convolution to align the size and channel number with that of the lower-level stages. These features are then added to the low-level features (i.e., input\_1) that have been processed by channel attention. Here, spatial average pooling and spatial max pooling refer to pooling performed on each channel individually, sharing the same convolutional layers. The number of output channels of the channel attention module is consistent with that of the features from the lower-level stage. Following ConvNeXt \cite{RN907}, the expansion factor $\mathit{k}$ is set to 4. Subsequently, the output of the channel attention module is fed into multi-scale spatial attention. Inspired by CSWin \cite{RN910}, we design multi-scale cross-shaped convolutional kernels within the spatial attention to capture a broader range of contextual information with as few parameters as possible. The channel average pooling and channel max pooling refer to pooling along the channel axis. For the side outputs, features from the lower-level stages are first aligned in size and channel number via downsampling with convolution, and then the attention operation is conducted on the inputs. The channel number of the final output is set to match that of the features from the higher-level stage.
        
    \subsection{Edge fusion module (EFM)}
        The human visual system exhibits shape bias in object recognition \cite{RN887}. Accordingly, the edge detection task is introduced to provide shape constraints. To achieve edge prediction, we design a simple decoder composed of the EFM. Moreover, we utilize the decoder of the frozen LVP-Net \cite{RN747} pre-trained on the BSDS-VOC dataset \cite{RN627,RN502} to generate pseudo labels for edge prediction. As illustrated in Fig. \ref{fig4}\textcolor{red}{c}, inputs with different resolutions are individually processed through a convolutional layer, LN, and GELU activation \cite{RN676}. The lower-resolution input is then upsampled using bilinear interpolation and added to the other inputs. After several EFMs, an edge prediction map is generated via bilinear upsampling, a 3$\times$3 convolutional layer, and a sigmoid function in sequence.

    \subsection{Network training}
        To simultaneously optimize the edge detection task and the classification task, we define a multi-task learning loss function $L_{total}$ as follows:
        \begin{equation}
            L_{total}=L_{cls}+\gamma \cdot L_{edge},
        \end{equation}
        where $L_{cls}$ and $L_{edge}$ represent the classification loss and the edge loss, respectively. $\gamma$ is the fusion weight. In the ablation study, we carefully assess the impact of $\gamma$ on network performance in Table \ref{table5}.

        The classification loss $L_{cls}$ is defined as multi-class cross-entropy, which is formulated as
        \begin{equation}
            L_{cls}\left( P_{cls},Y_{cls} \right) =-\frac{1}{N}\sum_{i=0}^{N-1}{\sum_{j=0}^{M-1}{Y_{cls}^{ij}\log \left( P_{cls}^{ij} \right)}},
        \end{equation}
        where $Y_{cls}$ and $P_{cls}$ is the class labels and predictions, respectively. The shape of $P_{cls}$ and $Y_{cls}$ is $N\times$$M$, where $N$ and $M$ represents the sample size of a batch and the total classes, respectively. $Y_{cls}^{ij}$ is the $\left( i,j \right) ^{th}$ label of $Y_{cls}$, while $P_{cls}^{ij}$ is the $\left( i,j \right) ^{th}$ probability of $P_{cls}$.

        Due to the highly imbalanced distribution of positive and negative samples in the edge maps, we employ the class-balanced cross-entropy loss function proposed in HED \cite{RN481} as our edge loss. Following HED and LVP-Net \cite{RN747}, for an edge pseudo label $Y_{edge}=\left\{ y_j,j=1,...,\left| Y_{edg e} \right| \right\}$, $y_j\in \left[ 0,1 \right]$, we fist define the edge positive sample set and the edge negative sample set as $Y^+=\left\{ y_j,y_j>\eta \right\}$ and $Y^-=\left\{ y_j,y_j=0 \right\}$, respectively. $\eta$ is a threshold used to exclude ambiguous edge pixels. The other pixels are neglected. Given an edge prediction map $P_{edge}=\left\{ p_j,j=1,...,\left| P_{edge} \right| \right\}$, $p_j\in \left[ 0,1 \right]$, the edge loss function is formulated as
        \begin{equation}
            L_{edge}\left( P_{edge},Y_{edge} \right) =-\alpha \sum_{j\in Y^-}{\log \left( 1-p_j \right)}-\beta \sum_{j\in Y^+}{\log \left( p_j \right)},
        \end{equation}
        where
        \begin{equation}
            \alpha =\lambda \cdot \frac{\left| Y^+ \right|}{\left| Y^+ \right|+\left| Y^- \right|},
        \end{equation}

        \begin{equation}
            \beta =\frac{\left| Y^- \right|}{\left| Y^+ \right|+\left| Y^- \right|}.
        \end{equation}

        Here, $p_j$ is an edge probability value computed using the sigmoid function at the $j^{th}$ pixel. $\alpha$ and $\beta$ are the weights to control the balance between positive and negative samples. $\lambda$ is an empirical coefficient to adjust the weights of edge and non-edge loss.

\section{Experiments}
    \subsection{Datasets}
\begin{figure}[t]
            \centering
            \includegraphics[width=83mm]{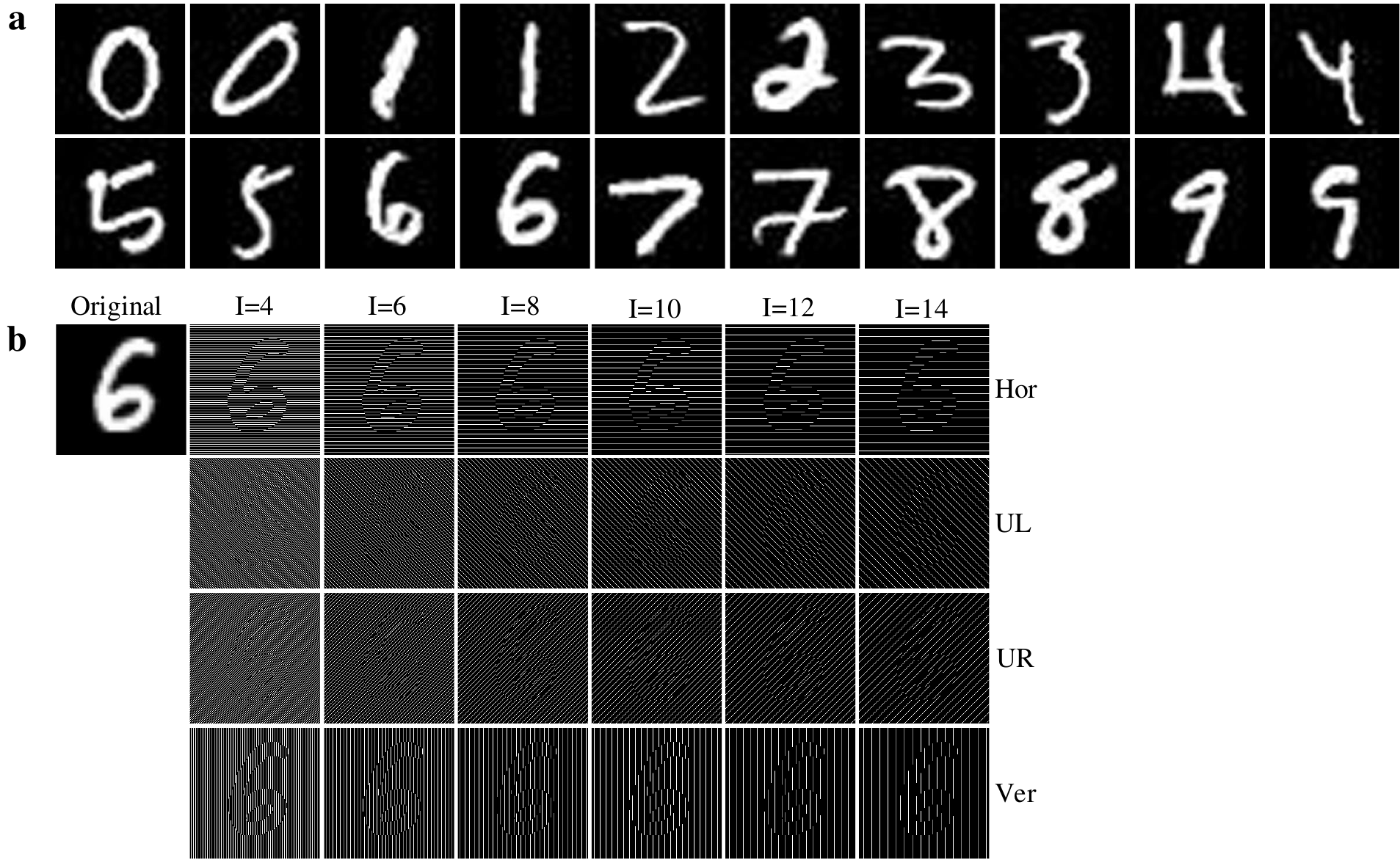}
            \caption{Examples in the MNIST and AG-MNIST test sets. \textbf{a.} Demonstration of some original handwritten digit images in the MNIST test set totaling ten categories, with two images per category. \textbf{b.} Demonstration of the abutting grating illusion images in the AG-MNIST test set, generated with four orientation gratings. "I" denotes the pixel interval between grating lines. "Hor", "UL", "UR", and "Ver" represent the illusory contour generated via horizontal, diagonal from upper left to lower right, diagonal from upper right to lower left, and vertical abutting gratings, respectively.}
            \label{fig5}
        \end{figure}

        We evaluated the capability of perceiving abutting grating illusory contours of various DNN models on the AG-MNIST \cite{RN756} and AG-Fashion-MNIST test sets. MNIST \cite{RN808} is a well-known handwritten digit image dataset that contains 60,000 training images and 10,000 testing images across ten categories. The AG-MNIST test set was constructed based on the abutting grating illusion and the MNIST test set. In this paper, the AG-MNIST test set comprises various types of images, which are generated by the abutting gratings in four orientations with six different pixel intervals between grating lines. AG-MNIST shares the same categories as MNIST. Following Fan et al. \cite{RN756}, since the small size of the original MNIST images, the AG-MNIST test set upsamples the original images to 224$\times$224 to retain more details. Fig. \ref{fig5} shows some examples.

\begin{figure}[t]
            \centering
            \includegraphics[width=83mm]{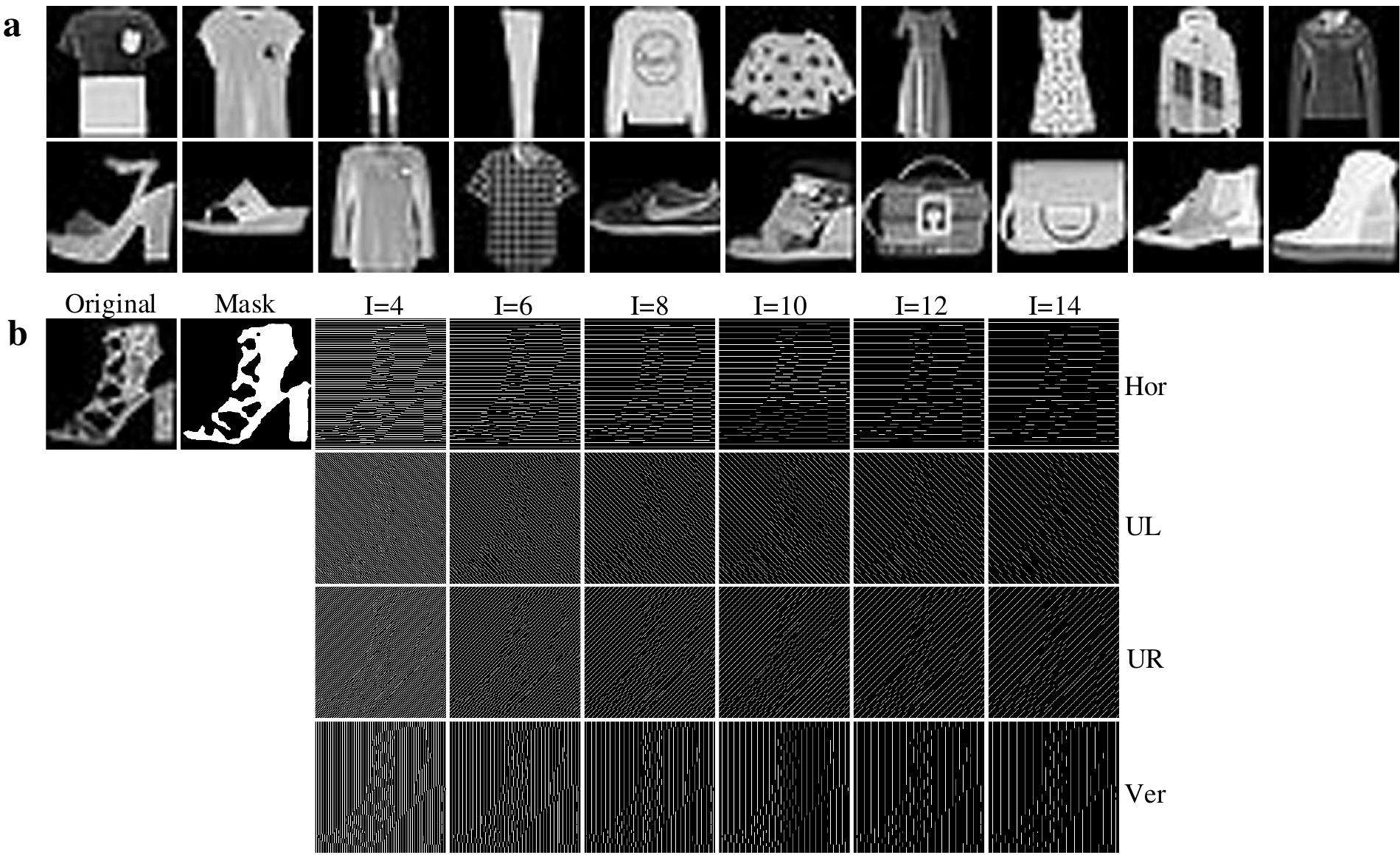}
            \caption{Examples in the Fashion-MNIST and AG-Fashion-MNIST test sets. \textbf{a.} Demonstration of some original fashion product images in the Fashion-MNIST test set totaling ten categories, with two images per category. \textbf{b.} Demonstration of the abutting grating illusion images in the AG-Fashion-MNIST test set, generated with four orientation gratings. "I" denotes the pixel interval between grating lines. "Hor", "UL", "UR", and "Ver" represent the illusory contour generated via horizontal, diagonal from upper left to lower right, diagonal from upper right to lower left, and vertical abutting gratings, respectively.}
            \label{fig6}
        \end{figure}

        To further evaluate the perception of abutting grating illusory contours, we transformed the more challenging Fashion-MNIST test set \cite{RN920} into the abutting grating illusory images. The Fashion-MNIST dataset comprises ten categories totaling 70,000 grayscale images of fashion products, with 7,000 images per category. The training set and test set consist of 60,000 and 10,000 images, respectively. We first extracted the masks from the Fashion-MNIST test set via the Segment Anything Model (SAM) \cite{RN922} and manually removed the inaccurate masks. For mutual complementarity, the Fashion-MNIST test set was also binarized by the OTSU method \cite{RN923} and the poorly binarized images were manually deleted. Subsequently, the two groups of images were mixed together to preserve as much of the test images as possible. Finally, the abutting grating distortion method \cite{RN756} was applied to the binarized images to generate the illusory images. The resulting test set is named the AG-Fashion-MNIST test set, which contains 8,916 abutting grating images. Fig. \ref{fig6} exhibits some examples.

        With the AG-MNIST and AG-Fashion-MNIST test sets, we can directly and quantitatively assess the abutting grating illusory contour perception ability of networks without the need for model fine-tuning or modifications to the classifier. Moreover, the AG-Fashion-MNIST test set is particularly suitable for researchers with limited computational resources due to its small scale.

    \subsection{Implementation details}
        In this study, we trained eight models from scratch on the MNIST and Fashion-MNIST training set, including VGG16 \cite{RN490}, ResNet18 \cite{RN608}, ResNet101 \cite{RN608}, ViT-L-16 \cite{RN745}, Swin-B \cite{RN809}, ConvNeXt-B \cite{RN907}, MambaOut-B \cite{RN810}, and our proposed ICPNet. All models were evaluated on both the original and AG test sets. Our network was initialized from zero-mean Gaussian distribution with a standard deviation of 0.02. For the edge detection task, following previous work \cite{RN481,RN747}, the threshold $\eta$ and the coefficient $\lambda$ were set to 0.2 and 1, respectively.

        For fairness, we maintained consistent hyperparameters as much as possible. To stabilize the training for ConvNeXt-B and MambaOut-B, we employed the AdamW \cite{RN912} optimizer with the initial learning rates of 1$\times10^{-3}$ and 4$\times10^{-4}$, respectively. Our ICPNet was trained using SGD with an initial learning rate of 1$\times10^{-3}$ to facilitate the convergence of both loss functions. The remaining models were optimized using SGD with an initial learning rate of 1$\times10^{-2}$. The momentum and batch size were set to 0.9 and 40, respectively. We adopted a standard input resolution of 224$\times$224. Following Fan et al. \cite{RN756}, all models were trained for 100 epochs, with the best-performing model on the original test sets selected for subsequent evaluation. Aside from random horizontal flipping, no additional tricks were applied, aiming to focus solely on the network architecture. All experiments were conducted on an NVIDIA 3090 GPU with 24GB of memory.

    \subsection{Ablation study}
        \begin{table*}[t]
            \centering
            \caption{Ablation experiment groups.}
            \setlength{\tabcolsep}{2pt}
            \renewcommand{\arraystretch}{1.2}
            \begin{tabular}{ccccccllll}
            \cline{1-6}
            Experiments & RF size & Feedback & MFP & FIAM & EFM &  &  &  &  \\ \cline{1-6}
            Group 1     & $\times$       & $\times$        & $\times$   & $\times$    & $\times$   &  &  &  &  \\
            Group 2     & $\surd$       & $\times$        & $\times$   & $\times$    & $\times$   &  &  &  &  \\
            Group 3     & $\surd$       & $\surd$        & $\times$   & $\times$    & $\times$   &  &  &  &  \\
            Group 4     & $\surd$       & $\surd$        & $\surd$   & $\times$    & $\times$   &  &  &  &  \\
            Group 5     & $\surd$       & $\surd$        & $\surd$   & $\surd$    & $\times$   &  &  &  &  \\
            Group 6     & $\surd$       & $\surd$        & $\surd$   & $\surd$    & $\surd$   &  &  &  &  \\ \cline{1-6}
            \end{tabular}
            \label{table2}
        \end{table*}

        \begin{table*}[t]
            \newcommand{\tabincell}[2]{\begin{tabular}{@{}#1@{}}#2\end{tabular}}
            \centering
            \caption{Ablation experiments of RF size, feedback, MFP, FIAM, and EFM. The first row of results is obtained on the original MNIST test set. Bold denotes the best top-1 accuracy.}
            \setlength{\tabcolsep}{2pt}
            \renewcommand{\arraystretch}{1.}
            \begin{tabular}{cccccccc}
                \hline
                \multicolumn{2}{l}{Dataset}         & \tabincell{c}{Group 1\\(Top-1 Acc.(\%))} & \tabincell{c}{Group 2\\(Top-1 Acc.(\%))} & \tabincell{c}{Group 3\\(Top-1 Acc.(\%))} & \tabincell{c}{Group 4\\(Top-1 Acc.(\%))} & \tabincell{c}{Group 5\\(Top-1 Acc.(\%))} & \tabincell{c}{Group 6\\(Top-1 Acc.(\%))} \\ \hline
                \multicolumn{2}{c}{\tabincell{c}{MNIST\\(Original)}} & 99.58                  & 99.45                  & 99.55                  & 99.49                  & 99.62                    & 99.49                    \\ \hline
                \multirow{4}{*}{I=4}      & Hor     & 28.37                  & 34.35                  & 49.02                  & 27.80                  & 31.36                    & \textbf{81.27}           \\
                                            & UL      & 17.97                  & 19.18                  & 19.09                  & 11.59                  & 18.75                    & \textbf{77.72}           \\
                                            & UR      & 24.52                  & 19.36                  & 16.65                  & 18.25                  & 26.87                    & \textbf{70.94}           \\
                                            & Ver     & 19.17                  & 23.16                  & 22.56                  & 11.20                  & 25.17                    & \textbf{67.43}           \\ \hline
                \multirow{4}{*}{I=6}      & Hor     & 18.06                  & 57.09                  & 49.70                  & 97.02                  & \textbf{98.54}           & 98.15                    \\
                                            & UL      & 13.78                  & 19.52                  & 40.89                  & 97.93                  & \textbf{98.85}           & 98.73                    \\
                                            & UR      & 14.78                  & 18.45                  & 34.76                  & 98.25                  & \textbf{98.91}           & 98.81                    \\
                                            & Ver     & 19.44                  & 47.48                  & 48.43                  & 98.01                  & \textbf{98.61}           & 98.50                    \\ \hline
                \multirow{4}{*}{I=8}      & Hor     & 29.88                  & 26.96                  & 20.46                  & 26.50                  & 37.69                    & \textbf{42.46}           \\
                                            & UL      & 13.13                  & 16.66                  & 24.98                  & 23.37                  & 32.02                    & \textbf{58.22}           \\
                                            & UR      & 20.54                  & 14.28                  & 23.32                  & 24.25                  & 38.53                    & \textbf{49.48}           \\
                                            & Ver     & \textbf{62.51}         & 23.19                  & 42.13                  & 20.39                  & 41.59                    & 35.22                    \\ \hline
                \multirow{4}{*}{I=10}     & Hor     & 15.42                  & 21.48                  & 17.93                  & 82.65                  & \textbf{94.97}           & 94.33                    \\
                                            & UL      & 19.34                  & 15.01                  & 38.11                  & 93.25                  & \textbf{98.46}           & 97.55                    \\
                                            & UR      & 14.75                  & 13.71                  & 25.58                  & 94.53                  & \textbf{98.47}           & 98.08                    \\
                                            & Ver     & 16.74                  & 17.06                  & 16.47                  & 83.97                  & 94.10                    & \textbf{95.90}           \\ \hline
                \multirow{4}{*}{I=12}     & Hor     & 13.10                  & 16.81                  & 12.44                  & 14.83                  & 10.45                    & \textbf{38.28}           \\
                                            & UL      & 11.67                  & 15.02                  & 22.82                  & 10.98                  & 24.51                    & \textbf{35.26}           \\
                                            & UR      & 19.23                  & 12.38                  & 18.56                  & 9.91                   & 21.84                    & \textbf{37.73}           \\
                                            & Ver     & 11.84                  & 12.69                  & 9.88                   & 11.29                  & 16.41                    & \textbf{27.89}           \\ \hline
                \multirow{4}{*}{I=14}     & Hor     & 11.27                  & 12.60                  & 22.47                  & 83.30                  & \textbf{94.83}           & 77.45                    \\
                                            & UL      & 14.44                  & 13.82                  & 41.10                  & 92.50                  & \textbf{98.49}           & 93.15                    \\
                                            & UR      & 14.99                  & 11.88                  & 31.63                  & 89.81                  & \textbf{98.13}           & 96.18                    \\
                                            & Ver     & 13.59                  & 15.58                  & 10.85                  & 68.18                  & \textbf{95.32}           & 86.09                    \\ \hline                
            \end{tabular}
            \label{table3}
        \end{table*}

        We conducted systematic ablation experiments to evaluate the model from multiple perspectives. Our network was trained on the MNIST training set and evaluated on both the MNIST and AG-MNIST test sets. As shown in Table \ref{table2}, for clarity, we first divided the experiments into six groups, including receptive field (RF) size, feedback interaction, MFP, FIAM, and EFM. The detailed quantitative results are provided in Table \ref{table3}.
        \begin{figure*}[t]
            \centering
            \includegraphics[width=170mm]{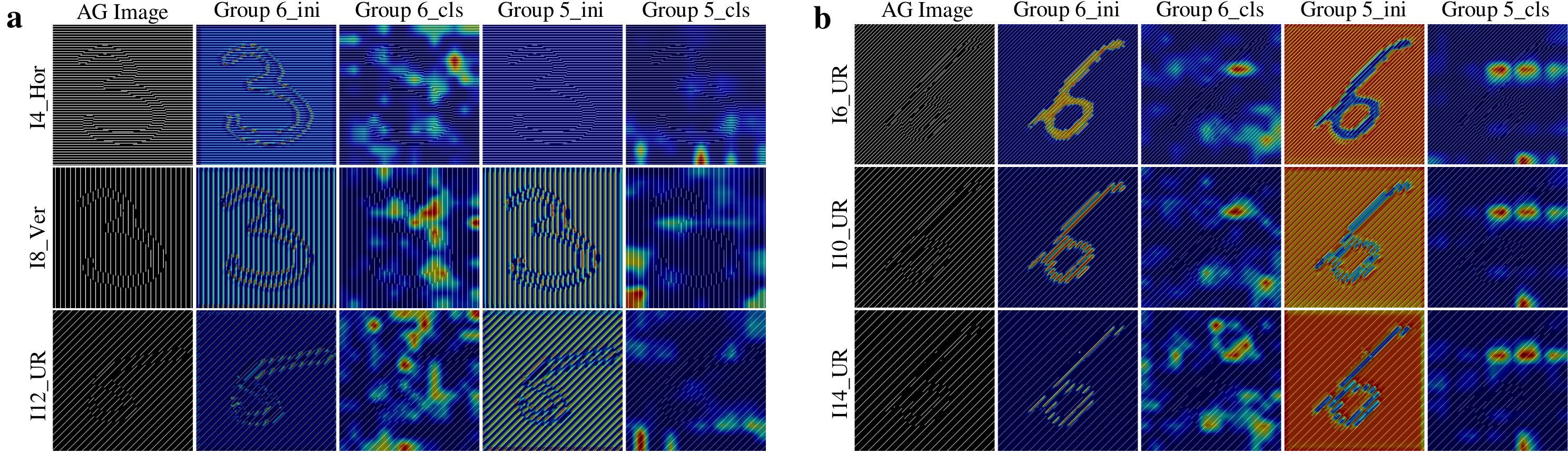}
            \caption{Visual explanations of the abutting grating illusion images. \textbf{a.} Visual explanations of the illusion images with intervals of 4, 8, and 12 generated by Grad-CAM. \textbf{b.} Visual explanations of the illusion images with intervals of 6, 10, and 14 generated by Grad-CAM. "ini" indicates the visualizations of the input features of the first stage, while "cls" represents the visualizations of the output features from the final convolutional layer of the network.}
            \label{fig7}
        \end{figure*}

        Overall, each component contributes to the improvement of the abutting grating illusory contours perception. In Group 1, all convolutional kernels across stages are uniformly set to 3×3, whereas Group 2 employs kernel configurations according to Table \ref{table1}. As shown in Table \ref{table3}, despite adopting a purely feedforward mode, the top-1 accuracy of Group 1 overall surpasses that of most of the methods reported in Table \ref{table6}. Compared with Group 1, Group 2 exhibits significant accuracy improvements at grating line intervals of 4 and 6. These results confirm the effectiveness of the foundational configurations. Upon incorporating feedback connections, the network exhibits further improvements, especially at the larger intervals of 10, 12, and 14. This finding indicates that feedback modulation with global information contributes to illusory contour perception, which is consistent with the discoveries of Pak et al. \cite{RN792} and Pan et al. \cite{RN767}. In Group 4, multi-scale RF significantly enhances accuracy at the intervals of 6, 10, and 14. Moreover, FIAM facilitates a comprehensive fusion of representations, boosting performance across most pixel intervals. Finally, edge detection based on the EFM provides shape constraints for the model. Although the results for the intervals of 6, 10, and 14 slightly decreased, the constraint markedly improves the top-1 accuracy at the intervals of 4, 8, and 12. Collectively, these results validate the complementary functionality of each module.

        \begin{figure}[t]
            \centering
            \includegraphics[width=83mm]{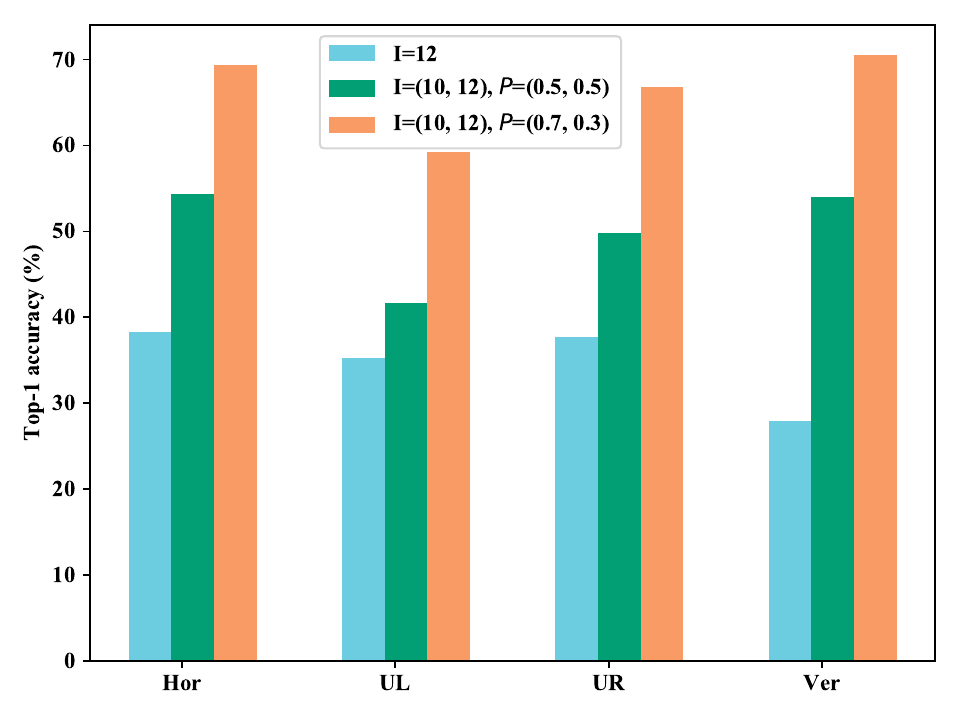}
            \caption{The top-1 accuracy on the abutting grating illusion images with randomly varying intervals. "I=(10, 12), $\mathit{P}$=(0.7, 0.3)" denotes the abutting gratings with intervals of 10 and 12 appear in each image with probabilities of 0.7 and 0.3, respectively.}
            \label{fig8}
        \end{figure}

        \begin{table}[t]
            \centering
            \caption{The effect of the number of grating lines on top‑1 accuracy. The first row corresponds to the deleted proportions of grating lines.}
            \setlength{\tabcolsep}{6pt}
            \renewcommand{\arraystretch}{1.1}
            \begin{tabular}{ccccccc}
            \hline
            \multicolumn{2}{c}{Dataset} & 0\%   & 10\%  & 20\%  & 40\%  & 80\%  \\ \hline
            \multirow{4}{*}{I=6}  & Hor & 98.15 & 95.74 & 91.62 & 71.18 & 22.69 \\
                                    & UL  & 98.73 & 97.91 & 96.29 & 87.82 & 36.62 \\
                                    & UR  & 98.81 & 98.24 & 96.95 & 90.66 & 36.40 \\
                                    & Ver & 98.50 & 96.74 & 94.62 & 82.86 & 34.05 \\ \hline
            \end{tabular}
            \label{table4}
        \end{table}

        For analyzing shape constraints, we employed Grad-CAM \cite{RN917} to visualize features for Group 5 and Group 6, as shown in Fig. \ref{fig7}. These examples were all correctly classified by the networks. First, it can be observed that the network indeed pays more attention to the foreground and the illusory contours after incorporating shape constraints, which aligns more closely with human intuition. Conversely, although the results of classification are correct, the visualizations for Group 5 tend to emphasize background informatiosn. Specifically, the visualizations of Group 6 for intervals of 6, 10, and 14 to some extent validate the complementarity between oriented illusory contour perception and non-oriented surface brightness filling-in \cite{RN918}.

        In conjunction with both quantitative and qualitative results, we posit that the reduced accuracy of ICPNet at certain intervals may stem from an intrinsic perceptual bias within the network, which impairs its ability to perceive abutting grating illusory contours at these intervals. In fact, the interval-dependent bias has been also reported in human experiments \cite{RN924}. To verify this hypothesis, we evaluated the performance of ICPNet on abutting grating images with randomly varying intervals. As shown in Fig. \ref{fig8}, compared to the grating images only with interval 12, increasing the proportion of interval 10 leads to the pronounced gains in top-1 accuracy, which proves our hypothesis.

        \begin{table*}[t]
            \newcommand{\tabincell}[2]{\begin{tabular}{@{}#1@{}}#2\end{tabular}}
            \centering
            \caption{The impact of the fusion weight $\gamma$ on ICPNet. The first row of results is obtained on the original MNIST test set. Bold denotes the best top-1 accuracy.}
            \setlength{\tabcolsep}{2pt}
            \renewcommand{\arraystretch}{1.}
            \begin{tabular}{ccccccc}
            \hline
            \multicolumn{2}{c}{Dataset}         & \tabincell{c}{$\gamma$=0.004\\Top-1 Acc.(\%)} & \tabincell{c}{$\gamma$=0.008\\Top-1 Acc.(\%)} & \tabincell{c}{$\gamma$=0.01\\Top-1 Acc.(\%)} & \tabincell{c}{$\gamma$=0.02\\Top-1 Acc.(\%)} & \tabincell{c}{$\gamma$=0.04\\Top-1 Acc.(\%)} \\ \hline
            \multicolumn{2}{c}{MNIST(Original)} & 99.49                  & 99.45                  & 99.49                 & 99.55                 & 99.28                 \\ \hline
            \multirow{4}{*}{I=4}      & Hor     & 29.08                  & 37.93                  & \textbf{81.27}                 & 60.34                 & 15.32                 \\
                                    & UL      & 63.81                  & 44.93                  & \textbf{77.72}                 & 44.13                 & 12.67                 \\
                                    & UR      & 53.85                  & 41.60                  & \textbf{70.94}                 & 55.79                 & 18.44                 \\
                                    & Ver     & 35.57                  & 35.98                  & \textbf{67.43}                 & 30.27                 & 13.32                 \\ \hline
            \multirow{4}{*}{I=6}      & Hor     & 97.99                  & \textbf{98.29}                  & 98.15                 & 97.37                 & 97.29                 \\
                                    & UL      & 98.44                  & 98.49                  & \textbf{98.73}                 & 97.23                 & 94.16                 \\
                                    & UR      & \textbf{98.85}                  & 98.35                  & 98.81                 & 97.74                 & 91.79                 \\
                                    & Ver     & 97.82                  & 98.35                  & \textbf{98.50}                 & 97.74                 & 92.06                 \\ \hline
            \multirow{4}{*}{I=8}      & Hor     & 34.29                  & 37.37                  & \textbf{42.46}                 & 36.94                 & 29.38                 \\
                                    & UL      & 41.42                  & \textbf{59.47}                  & 58.22                 & 36.43                 & 28.10                 \\
                                    & UR      & 26.81                  & 32.54                  & \textbf{49.48}                 & 29.56                 & 28.02                 \\
                                    & Ver     & 25.87                  & \textbf{36.17}                  & 35.22                 & 28.49                 & 23.61                 \\ \hline
            \multirow{4}{*}{I=10}     & Hor     & 88.34                  & 89.03                  & \textbf{94.33}                 & 91.26                 & 66.67                 \\
                                    & UL      & 97.35                  & 97.10                  & \textbf{97.55}                 & 96.65                 & 89.82                 \\
                                    & UR      & \textbf{98.15}                  & 97.04                  & 98.08                 & 98.09                 & 83.51                 \\
                                    & Ver     & 86.96                  & 94.01                  & \textbf{95.90}                 & 94.17                 & 84.17                 \\ \hline
            \multirow{4}{*}{I=12}     & Hor     & 27.36                  & 23.27                  & \textbf{38.28}                 & 25.29                 & 25.34                 \\
                                    & UL      & \textbf{38.51}                  & 25.67                  & 35.26                 & 33.56                 & 20.32                 \\
                                    & UR      & 26.21                  & 20.32                  & \textbf{37.73}                 & 22.65                 & 10.84                 \\
                                    & Ver     & 19.29                  & 12.97                  & \textbf{27.89}                 & 18.49                 & 11.07                 \\ \hline
            \multirow{4}{*}{I=14}     & Hor     & 75.92                  & 69.90                  & \textbf{77.45}                 & 71.22                 & 54.41                 \\
                                    & UL      & \textbf{96.15}                  & 93.25                  & 93.15                 & 95.71                 & 85.28                 \\
                                    & UR      & \textbf{96.31}                  & 95.99                  & 96.18                 & 94.80                 & 73.84                 \\
                                    & Ver     & 81.65                  & 82.95                  & \textbf{86.09}                 & 80.07                 & 55.72                 \\ \hline
            \end{tabular}
            \label{table5}
        \end{table*}

        Moreover, to examine the effect of the number of grating lines on top-1 accuracy, we conducted experiments at the grating images of interval 6, where we randomly deleted varying proportions of grating lines, as shown in Table \ref{table4}. The results reveal that the accuracy of ICPNet consistently declines as grating lines are removed, with a steep drop observed at the proportions of 40\% and 80\%. These results align with previous observations that sparser grating lines typically accompany weaker illusory contour perception strength \cite{RN924,RN786}.

        Finally, we evaluated the impact of the fusion weight $\gamma$ of the multi-task loss function on ICPNet. As shown in Table \ref{table5}, the suboptimal fusion leads to a significant degradation in the top-1 accuracy of abutting grating illusory images. In this study, the fusion weight $\gamma$ is set to 0.01.

    \subsection{Comparison with state-of-the-art methods}
        \textbf{Performance comparison on AG-MNIST.} We compare ICPNet with several representative DNN models, including VGG16 \cite{RN490}, ResNet18 \cite{RN608}, ResNet101 \cite{RN608}, ViT-L-16 \cite{RN745}, Swin-B \cite{RN809}, ConvNeXt-B \cite{RN907}, and MambaOut-B \cite{RN810}. All methods were trained from scratch on the MNIST training set \cite{RN808} and then directly evaluated on the AG-MNIST test set \cite{RN756}. The ICPNet variant with only feedback connections (i.e., Group 3 in Table \ref{table3}) is selected as the baseline. The detailed quantitative results are presented in Table \ref{table6}.
        \begin{table*}[!h]
            \newcommand{\tabincell}[2]{\begin{tabular}{@{}#1@{}}#2\end{tabular}}
            \centering
            \caption{Quantitative comparison on MNIST and AG-MNIST test sets. The human data derives from the work of Fan et al. \cite{RN756}. The first row of results is obtained on the original MNIST test set. Group 3 in Table \ref{table3} is selected as the baseline. The best two results are bolded and highlighted in \textbf{\textcolor{red}{red}} and \textbf{\textcolor{blue}{blue}}, respectively.}
            \setlength{\tabcolsep}{1.5pt}
            \renewcommand{\arraystretch}{1.2}
            \begin{tabular}{cccccccccccc}
            \hline
            \multicolumn{2}{l}{\multirow{2}{*}{Dataset}}                                         & \multirow{2}{*}{\tabincell{c}{Human \\(Top-1 Acc. (\%))}} & \multicolumn{9}{c}{Model (Top-1 Acc. (\%))}                                                                                        \\ \cline{4-12} 
            \multicolumn{2}{l}{}                                                                 &                                        & Baseline       & VGG16          & ResNet18 & ResNet101      & \tabincell{c}{ViT-L-16} & Swin-B         & \tabincell{c}{ConvNeXt\\-B} & \tabincell{c}{MambaOut\\-B} & \tabincell{c}{Ours\\(ICPNet)}   \\ \hline
            \multicolumn{2}{l}{\begin{tabular}[c]{@{}l@{}}MNIST\\(Original)\end{tabular}} & --                                     & 99.55          & 99.41          & 99.70    & 99.72          & 97.23    & 98.62          & 99.45      & 99.37      & 99.49          \\ \hline
            \multirow{4}{*}{I=4}                               & Hor                             & 98.33                                  & \textbf{\textcolor{blue}{49.02}} & 17.18          & 9.80     & 9.74           & 9.06     & 7.95           & 11.99      & 14.02      & \textbf{\textcolor{red}{81.27}} \\
                                                            & UL                              & 98.67                                  & 19.09          & 17.22          & 9.80     & 9.74           & 9.80     & \textbf{\textcolor{red}{95.74}} & 8.74       & 11.49      & \textbf{\textcolor{blue}{77.72}} \\
                                                            & UR                              & 98.67                                  & 16.65          & \textbf{\textcolor{blue}{16.72}} & 9.80     & 9.74           & 8.63     & 11.50          & 10.79      & 12.77      & \textbf{\textcolor{red}{70.94}} \\
                                                            & Ver                             & 97.67                                  & \textbf{\textcolor{blue}{22.56}} & 6.22           & 9.80     & 9.74           & 9.48     & 9.74           & 10.39      & 8.24       & \textbf{\textcolor{red}{67.43}} \\ \hline
            \multirow{4}{*}{I=6}                               & Hor                             & --                                     & \textbf{\textcolor{blue}{49.70}} & 14.40          & 9.80     & 9.74           & 8.34     & 9.74           & 10.69      & 9.32       & \textbf{\textcolor{red}{98.15}} \\
                                                            & UL                              & --                                     & \textbf{\textcolor{blue}{40.89}} & 16.97          & 9.80     & 9.74           & 9.81     & 9.74           & 19.07      & 8.81       & \textbf{\textcolor{red}{98.73}} \\
                                                            & UR                              & --                                     & \textbf{\textcolor{blue}{34.76}} & 14.53          & 9.80     & 9.74           & 9.82     & 9.74           & 9.71       & 24.69      & \textbf{\textcolor{red}{98.81}} \\
                                                            & Ver                             & --                                     & \textbf{\textcolor{blue}{48.43}} & 6.50           & 9.80     & 9.74           & 9.78     & 10.20          & 9.95       & 6.89       & \textbf{\textcolor{red}{98.50}} \\ \hline
            \multirow{4}{*}{I=8}                               & Hor                             & 98.67                                  & \textbf{\textcolor{blue}{20.46}} & 17.53          & 9.80     & 9.74           & 9.70     & 9.80           & 9.35       & 11.35      & \textbf{\textcolor{red}{42.46}} \\
                                                            & UL                              & 96.00                                  & \textbf{\textcolor{blue}{24.98}} & 12.29          & 9.80     & 9.74           & 10.59    & 17.68          & 10.52      & 12.24      & \textbf{\textcolor{red}{58.22}} \\
                                                            & UR                              & 97.67                                  & \textbf{\textcolor{blue}{23.32}} & 13.61          & 9.80     & 9.74           & 10.15    & 9.74           & 9.29       & 18.22      & \textbf{\textcolor{red}{49.48}} \\
                                                            & Ver                             & 94.67                                  & \textbf{\textcolor{red}{42.13}} & 8.58           & 9.80     & 9.74           & 9.06     & 9.77           & 20.85      & 11.35      & \textbf{\textcolor{blue}{35.22}} \\ \hline
            \multirow{4}{*}{I=10}                              & Hor                             & --                                     & 17.93          & \textbf{\textcolor{blue}{20.31}} & 9.80     & 9.74           & 7.00     & 9.74           & 8.93       & 10.95      & \textbf{\textcolor{red}{94.33}} \\
                                                            & UL                              & --                                     & \textbf{\textcolor{blue}{38.11}} & 15.34          & 9.80     & 9.74           & 9.67     & 9.74           & 14.90      & 11.36      & \textbf{\textcolor{red}{97.55}} \\
                                                            & UR                              & --                                     & \textbf{\textcolor{blue}{25.58}} & 15.20          & 9.80     & 9.74           & 10.05    & 9.79           & 11.00      & 11.48      & \textbf{\textcolor{red}{98.08}} \\
                                                            & Ver                             & --                                     & \textbf{\textcolor{blue}{16.47}} & 8.42           & 9.80     & 9.69           & 9.83     & 9.82           & 12.05      & 16.05      & \textbf{\textcolor{red}{95.90}} \\ \hline
            \multirow{4}{*}{I=12}                              & Hor                             & --                                     & 12.44          & \textbf{\textcolor{blue}{19.33}} & 11.35    & 18.72          & 10.09    & 9.73           & 9.45       & 11.35      & \textbf{\textcolor{red}{38.28}} \\
                                                            & UL                              & --                                     & 22.82          & 17.96          & 9.80     & 9.74           & 9.75     & 9.79           & \textbf{\textcolor{blue}{32.53}}      & 11.35      & \textbf{\textcolor{red}{35.26}} \\
                                                            & UR                              & --                                     & \textbf{\textcolor{blue}{18.56}} & 14.86          & 9.80     & 9.74           & 10.51    & 9.74           & 10.32      & 11.36      & \textbf{\textcolor{red}{37.73}} \\
                                                            & Ver                             & --                                     & 9.88           & 7.55           & 11.35    & \textbf{\textcolor{red}{44.59}} & 15.83    & 9.82           & 23.62      & 11.35      & \textbf{\textcolor{blue}{27.89}} \\ \hline
            \multirow{4}{*}{I=14}                              & Hor                             & --                                     & \textbf{\textcolor{blue}{22.47}} & 13.32          & 9.80     & 9.73           & 10.09    & 9.76           & 9.65       & 8.12       & \textbf{\textcolor{red}{77.45}} \\
                                                            & UL                              & --                                     & \textbf{\textcolor{blue}{41.10}} & 17.90          & 19.27    & 20.31          & 9.83     & 11.26          & 10.25      & 11.35      & \textbf{\textcolor{red}{93.15}} \\
                                                            & UR                              & --                                     & \textbf{\textcolor{blue}{31.63}} & 13.22          & 11.36    & 15.16          & 9.61     & 10.57          & 8.86       & 11.35      & \textbf{\textcolor{red}{96.18}} \\
                                                        & Ver                             & --                                     & 10.85          & 7.94           & 9.79     & 9.94           & 10.10    & 9.83           & 9.80       & \textbf{\textcolor{blue}{14.49}}      & \textbf{\textcolor{red}{86.09}} \\ \hline
            \end{tabular}
            \label{table6}
        \end{table*}

        \begin{figure*}[!h]
            \centering
            \includegraphics[width=150mm]{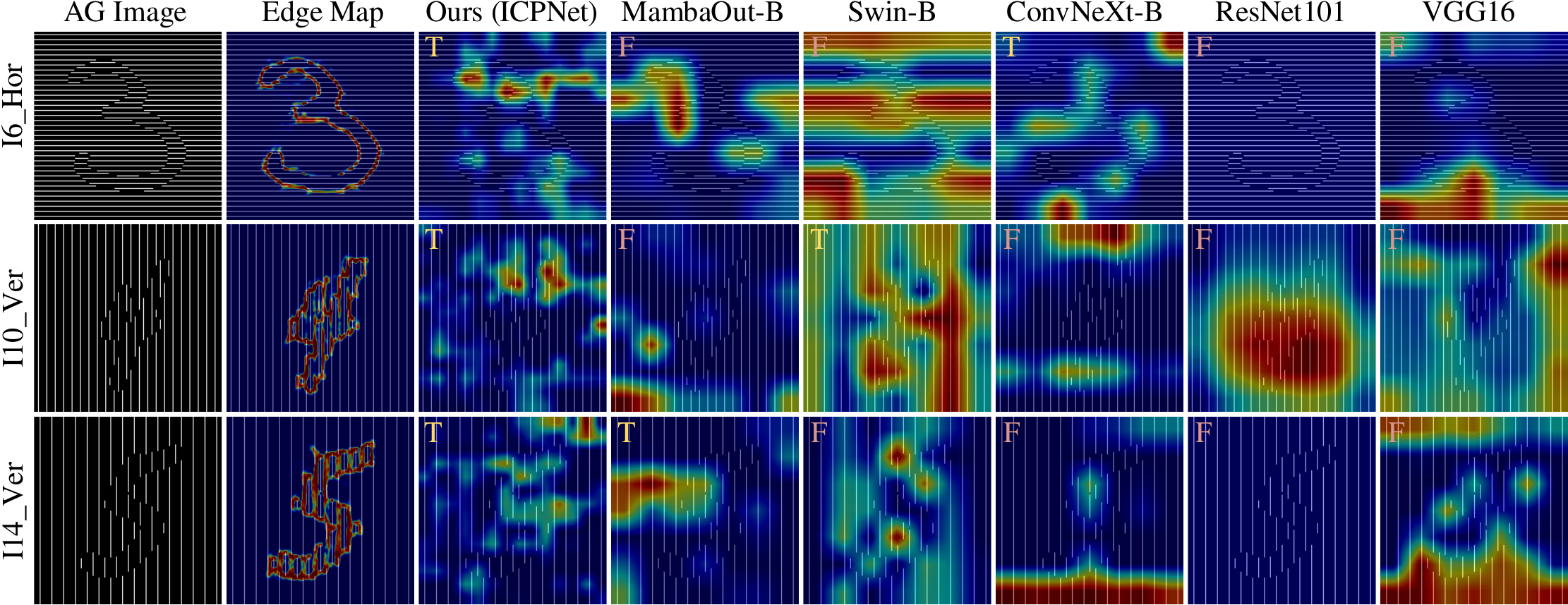}
            \caption{Visual explanations generated by Grad-CAM on the AG-MNIST test set. "Edge Map" refers to the edge prediction map produced by ICPNet. The remaining figures display the Grad-CAM visualizations from the output features of the final convolutional layer for each method. "I", "T", and "F" denote the pixel interval between grating lines, the correct, and incorrect prediction category, respectively. "Hor" and "Ver" represent the illusory contours generated via horizontal and vertical abutting gratings, respectively.}
            \label{fig9}
        \end{figure*}

        The quantitative results indicate that our approach exhibits a significant ability to perceive abutting grating illusory contours and our top-1 accuracy substantially outperforms that of state-of-the-art methods across most pixel intervals. Specifically, ICPNet achieves the average top-1 accuracy of 89.39\% at intervals of 4, 6, 10, and 14. Although the accuracy at intervals of 8 and 12 is relatively lower, it is still superior to most methods. We think the relative degradation of our model at these two intervals may stem from an intrinsic perceptual bias, as illustrated in Fig. \ref{fig8}. Notably, the top-1 accuracy of the baseline exceeds that of other methods on most subsets, underscoring the effectiveness of our feedback-based architecture. On the other hand, Swin attains a top-1 accuracy of 95.74\% for the UL gratings of interval 4, which may suggest an advantage of Transformers in long-range perception. However, on the other subsets, Swin performed near the random level. Overall, these results highlight the superiority of ICPNet and indicate that existing methods generally lack the capability of perceiving abutting grating illusory contours, which is consistent with the findings of Fan et al. \cite{RN756}.

        We also utilized Grad-CAM \cite{RN917} to generate visual explanations, as shown in Fig. \ref{fig9}. The edge maps indicate that our approach predicts the abutting grating illusory contours effectively. Furthermore, the class activation maps reveal that ICPNet focuses more on the foreground overall, which aligns with human intuition (see Fig. \ref{fig7} for more visual explanations of shape constraints). For example, ICPNet correctly classifies the digit "4" by concentrating on its two top areas, whereas Swin exhibits a more dispersive pattern and pays more attention to the background.

        \begin{table*}[t]
            \newcommand{\tabincell}[2]{\begin{tabular}{@{}#1@{}}#2\end{tabular}}
            \centering
            \caption{Quantitative comparison on Fashion-MNIST and AG-Fashion-MNIST test sets. The first row of results is obtained on the original Fashion-MNIST test set. The best two results are bolded and highlighted in \textbf{\textcolor{red}{red}} and \textbf{\textcolor{blue}{blue}}, respectively.}
            \setlength{\tabcolsep}{2pt}
            \renewcommand{\arraystretch}{1.}
            \begin{tabular}{cccccccccc}
                \hline
                \multicolumn{2}{l}{{ }}                                                                 & \multicolumn{8}{c}{{ Model (Top-1 Acc. (\%))}}                                                                                                                                                                                                                                                      \\ \cline{3-10} 
                \multicolumn{2}{l}{\multirow{-2}{*}{Dataset}}                                        & VGG16          & ResNet18       & ResNet101      & ViT-L-16 & Swin-B         & ConvNeXt-B     & MambaOut-B     & Ours(ICPNet)   \\ \hline
                \multicolumn{2}{l}{{ \begin{tabular}[c]{@{}l@{}}Fashion-MNIST\\(Original)\end{tabular}}} & 93.99          & 93.93          & 93.76          & 88.32    & 90.41          & 93.14          & 92.40          & 93.61          \\ \hline
                { }                                        & Hor                 & 10.13          & 10.34          & 11.04          & 10.24    & 12.65          & \textbf{\textcolor{blue}{19.30}} & 15.97          & \textbf{\textcolor{red}{44.48}} \\
                { }                                        & UL                  & 10.14          & \textbf{\textcolor{blue}{10.45}} & 10.14          & 8.58     & 9.43           & 10.08          & 11.20          & \textbf{\textcolor{red}{42.69}} \\
                { }                                        & UR                  & 10.14          & \textbf{\textcolor{blue}{10.46}} & 10.14          & 10.04    & 8.22           & 10.21          & 12.58          & \textbf{\textcolor{red}{42.25}} \\
                \multirow{-4}{*}{{I=4}}                   & Ver                 & 10.22          & 16.21          & 16.23          & 8.00     & 4.34           & \textbf{\textcolor{blue}{21.25}} & 15.28          & \textbf{\textcolor{red}{47.81}} \\ \hline
                { }                                        & Hor                 & \textbf{\textcolor{blue}{15.85}} & 15.66          & 13.92          & 10.23    & 5.18           & 10.14          & 15.93          & \textbf{\textcolor{red}{68.53}} \\
                { }                                        & UL                  & 10.14          & \textbf{\textcolor{blue}{17.74}} & 10.04          & 9.66     & 12.90          & 9.53           & 10.04          & \textbf{\textcolor{red}{77.15}} \\
                { }                                        & UR                  & 10.14          & 11.53          & \textbf{\textcolor{blue}{17.92}} & 10.14    & 12.63          & 10.16          & 16.07          & \textbf{\textcolor{red}{77.36}} \\
                \multirow{-4}{*}{{I=6}}                   & Ver                 & 14.75          & \textbf{\textcolor{blue}{24.20}} & 13.31          & 11.46    & 10.14          & 11.27          & 16.44          & \textbf{\textcolor{red}{73.83}} \\ \hline
                { }                                        & Hor                 & \textbf{\textcolor{blue}{21.02}} & 10.69          & 10.67          & 10.08    & 11.14          & 12.88          & 11.40          & \textbf{\textcolor{red}{48.04}} \\
                { }                                        & UL                  & 10.14          & 12.49          & 11.29          & 9.80     & \textbf{\textcolor{blue}{12.97}} & 9.37           & 11.05          & \textbf{\textcolor{red}{46.39}} \\
                { }                                        & UR                  & 13.29          & 12.63          & 11.01          & 11.77    & \textbf{\textcolor{blue}{17.15}} & 9.93           & 11.09          & \textbf{\textcolor{red}{42.90}} \\
                \multirow{-4}{*}{{I=8}}                   & Ver                 & \textbf{\textcolor{blue}{20.64}} & 18.29          & 8.65           & 3.53     & 3.60           & 16.67          & 19.64          & \textbf{\textcolor{red}{50.42}} \\ \hline
                { }                                        & Hor                 & \textbf{\textcolor{blue}{26.07}} & 9.85           & 14.39          & 7.69     & 2.21           & 14.38          & 13.44          & \textbf{\textcolor{red}{52.95}} \\
                { }                                        & UL                  & 10.04          & 10.66          & 10.74          & 6.71     & 10.14          & 12.11          & \textbf{\textcolor{blue}{15.53}} & \textbf{\textcolor{red}{68.51}} \\
                { }                                        & UR                  & 15.59          & 10.86          & 6.45           & 10.89    & 12.97          & 10.18          & \textbf{\textcolor{blue}{19.12}} & \textbf{\textcolor{red}{69.65}} \\
                \multirow{-4}{*}{{I=10}}                  & Ver                 & \textbf{\textcolor{blue}{22.80}} & 9.28           & 10.86          & 10.53    & 17.66          & 10.92          & 13.97          & \textbf{\textcolor{red}{38.75}} \\ \hline
                { }                                        & Hor                 & 28.93          & \textbf{\textcolor{red}{54.74}} & 27.32          & 12.98    & 6.71           & 10.92          & 20.82          & \textbf{\textcolor{blue}{34.75}} \\
                { }                                        & UL                  & 11.25          & 6.71           & 6.71           & 12.33    & 12.54          & \textbf{\textcolor{blue}{20.52}} & 17.07          & \textbf{\textcolor{red}{23.95}} \\
                { }                                        & UR                  & \textbf{\textcolor{blue}{18.29}} & 6.75           & 6.71           & 10.13    & 6.22           & 10.88          & 17.29          & \textbf{\textcolor{red}{18.93}} \\
                \multirow{-4}{*}{{I=12}}                  & Ver                 & \textbf{\textcolor{blue}{24.55}} & \textbf{\textcolor{red}{38.25}} & 9.70           & 10.74    & 9.99           & 12.64          & 23.33          & 13.95          \\ \hline
                { }                                        & Hor                 & \textbf{\textcolor{blue}{27.94}} & 10.14          & 10.14          & 10.14    & 6.73           & 10.03          & 11.59          & \textbf{\textcolor{red}{36.50}} \\
                { }                                        & UL                  & 14.86          & 6.77           & 6.71           & 10.14    & 10.08          & 12.91          & \textbf{\textcolor{blue}{15.06}} & \textbf{\textcolor{red}{59.34}} \\
                { }                                        & UR                  & 17.29          & 6.77           & 6.71           & 10.06    & 10.14          & 10.89          & \textbf{\textcolor{blue}{19.84}} & \textbf{\textcolor{red}{61.03}} \\
                \multirow{-4}{*}{{I=14}}                  & Ver                 & \textbf{\textcolor{blue}{18.30}} & 7.28           & 15.49          & 10.53    & 10.15          & 7.03           & 6.49           & \textbf{\textcolor{red}{34.05}} \\ \hline                
            \end{tabular}
            \label{table7}
        \end{table*}

        \begin{figure*}[t]
            \centering
            \includegraphics[width=150mm]{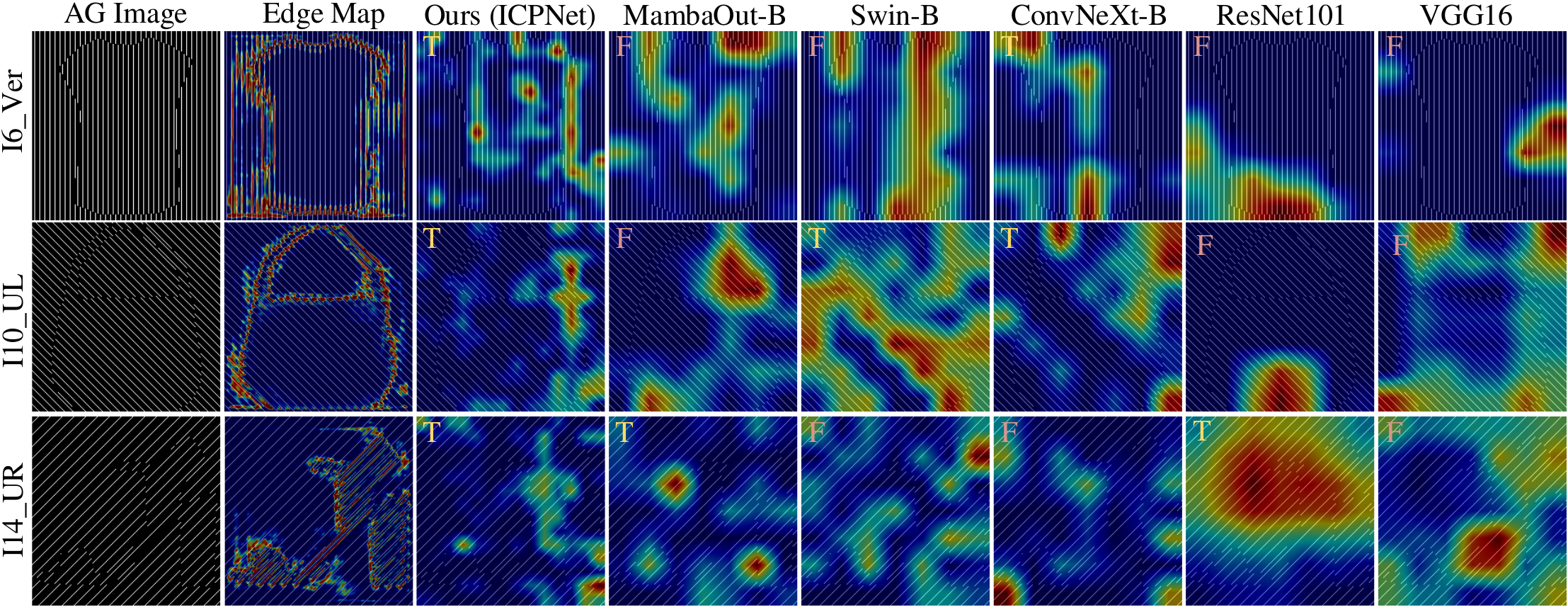}
            \caption{Visual explanations generated by Grad-CAM on the AG-Fashion-MNIST test set. "Edge Map" refers to the edge prediction map produced by ICPNet. The remaining figures display the Grad-CAM visualizations from the output features of the final convolutional layer for each method. "I", "T", and "F" denote the pixel interval between grating lines, the correct, and incorrect prediction category, respectively. "Ver", "UL", and "UR" represent the illusory contour generated via vertical, diagonal from upper left to lower right, and diagonal from upper right to lower left abutting gratings, respectively.}
            \label{fig10}
        \end{figure*}

        \textbf{Performance comparison on AG-Fashion-MNIST.} To further assess the perception of abutting grating illusory contours, we created the more challenging AG-Fashion-MNIST test set. We trained all networks from scratch on the Fashion-MNIST training set \cite{RN920} and conducted the evaluation on the AG-Fashion-MNIST test set. Table \ref{table7} presents the detailed quantitative results. Overall, the results indicate that ICPNet outperforms the state-of-the-art methods by a significant margin across most pixel intervals. Specifically, our method achieves approximately twice the top-1 accuracy of competing approaches at intervals of 4, 6, 8, 10, and 14, and even higher on some subsets. At interval 12, however, ICPNet shows only average performance, which is consistent with the observations from the AG-MNIST test set. Additionally, the performance on the AG-Fashion-MNIST test set is much lower than that on the AG-MNIST test set. We attribute this difference to the inherently higher classification difficulty of the Fashion-MNIST dataset, where objects exhibit more complex shapes and ambiguous category definitions compared to the simple digits in the AG-MNIST. Such discrepancies in performance also align with the established discoveries in human experiments \cite{RN756}. Critical evaluation reveals that, despite these challenges, ICPNet maintains a higher top-1 accuracy on most abutting grating illusory subsets, indicating a robust ability to perceive illusory contours.

        The visual explanations for some models generated by Grad-CAM \cite{RN917} are shown in Fig. \ref{fig10}. Compared with the edge maps of the AG-MNIST test set, the edge maps of the AG-Fashion-MNIST test set contain more background texture, which may derive from the inherent classification challenge of the dataset. Furthermore, the class activation maps reveal that our model captures multiple key regions for correct predictions. For example, ICPNet correctly classifies the sandal by concentrating on the toe, heel, and other parts, whereas Mambaout pays more attention to the background relatively.

\section{Discussion and conclusion}
    The abutting grating illusory contour is one of the most frequently used illusory figures for investigating shape perception and object recognition in the neuroscience field. Moreover, the abutting grating illusory contour perception represents a sophisticated and fundamental capability in biological visual systems, yet it is notably absent in DNNs, which are employed to simulate biological vision. In this study, inspired by the cortical structures and circuits of the visual systems, we construct a perceptual network with feedback connections, named illusory contour perception network (ICPNet). Extensive experimental results on the AG-MNIST and AG-Fashion-MNIST test sets demonstrate that, although a performance gap remains relative to human performance, ICPNet exhibits obviously conspicuous sensitivity to abutting grating illusory contours compared with other neural networks. This research underscores the critical role played by feedback modulation, attention mechanisms, multi-scale receptive fields, and shape bias in the perception of abutting grating illusory contours. While this study primarily focuses on the abutting grating illusion, we hope that ICPNet with hierarchical feedback connections will inspire the development of more robust networks from the perspective of structural innovation. Moreover, our results emphasize the necessity of fundamental cues (e.g., shape features) for machine perception under extreme distortion conditions. Consequently, the proposed multi-task learning framework holds the potential to inform visual model design in challenging tasks, such as camouflaged object detection \cite{RN925}, occlusion-aware segmentation \cite{RN927}, and underwater object detection \cite{RN926}.
    
    To sum up, illusory contour perception involves multiple cognitive and neural mechanisms. Beyond humans, this ability also serves as an anti-camouflage device in the natural world, enhancing the robustness in object recognition under degraded or occluded conditions for animals \cite{RN776}. In the era of rapidly evolving AI, the alignment of neural networks with human perception is increasingly crucial, as the ultimate goal of AI is to construct machines that think and perceive like humans \cite{RN885,RN823}. By leveraging visual mechanisms, our model attempts to narrow the gap between the behaviour of DNNs and human visual perception. Our approach presents a promising insight into integrating computer vision with biological vision. Future research will focus on augmenting the ability of model to detect various illusory contours and apply it to natural image understanding tasks.

\noindent\textbf{Acknowledgements} This study was supported by the STI2030-Major Projects (2022ZD0204600) and the National Natural Science Foundation of China (62476050).

{\small
	\bibliographystyle{IEEEtran}
	\bibliography{references}

@article{RN919,
   author = {Albright, Thomas D and Jessell, Thomas M and Kandel, Eric R and Posner, Michael I},
   title = {Neural science: a century of progress and the mysteries that remain},
   journal = {Cell},
   volume = {100},
   pages = {1-55},
   ISSN = {0092-8674},
   year = {2000},
   type = {Journal Article}
}

@article{RN627,
   author = {Arbelaez, Pablo and Maire, Michael and Fowlkes, Charless and Malik, Jitendra},
   title = {Contour detection and hierarchical image segmentation},
   journal = {IEEE Transactions on Pattern Analysis and Machine Intelligence},
   volume = {33},
   number = {5},
   pages = {898-916},
   ISSN = {0162-8828},
   year = {2010},
   type = {Journal Article}
}

@article{RN911,
   author = {Ba, Jimmy Lei and Kiros, Jamie Ryan and Hinton, Geoffrey E},
   title = {Layer normalization},
   journal = {arXiv preprint arXiv:1607.06450},
   year = {2016},
   type = {Journal Article}
}

@inproceedings{RN768,
   author = {Baker, Nicholas and Erlikhman, Gennady and Kellman, Philip J and Lu, Hongjing},
   title = {Deep Convolutional Networks do not Perceive Illusory Contours},
   booktitle = {Proceedings of the annual meeting of the cognitive science society},
   year = {2018},
   page = {1310-1315},
   volume = 40,
   type = {Conference Proceedings}
}

@article{RN925,
   author = {Bi, Hongbo and Zhang, Cong and Wang, Kang and Tong, Jinghui and Zheng, Feng},
   title = {Rethinking camouflaged object detection: Models and datasets},
   journal = {IEEE transactions on circuits and systems for video technology},
   volume = {32},
   number = {9},
   pages = {5708-5724},
   ISSN = {1051-8215},
   year = {2021},
   type = {Journal Article}
}

@article{RN754,
   author = {Bowers, Jeffrey S and Malhotra, Gaurav and Dujmović, Marin and Montero, Milton Llera and Tsvetkov, Christian and Biscione, Valerio and Puebla, Guillermo and Adolfi, Federico and Hummel, John E and Heaton, Rachel F},
   title = {Deep problems with neural network models of human vision},
   journal = {Behavioral and Brain Sciences},
   volume = {46},
   pages = {e385},
   ISSN = {0140-525X},
   year = {2023},
   type = {Journal Article}
}

@article{RN758,
   author = {Carbon, Claus-Christian},
   title = {Understanding human perception by human-made illusions},
   journal = {Frontiers in Human Neuroscience},
   volume = {8},
   pages = {566},
   ISSN = {1662-5161},
   year = {2014},
   type = {Journal Article}
}

@article{RN879,
   author = {Cox, Michele A and Schmid, Michael C and Peters, Andrew J and Saunders, Richard C and Leopold, David A and Maier, Alexander},
   title = {Receptive field focus of visual area V4 neurons determines responses to illusory surfaces},
   journal = {Proceedings of the National Academy of Sciences},
   volume = {110},
   number = {42},
   pages = {17095-17100},
   ISSN = {0027-8424},
   year = {2013},
   type = {Journal Article}
}

@article{RN808,
   author = {Deng, Li},
   title = {The mnist database of handwritten digit images for machine learning research},
   journal = {IEEE signal processing magazine},
   volume = {29},
   number = {6},
   pages = {141-142},
   ISSN = {1053-5888},
   year = {2012},
   type = {Journal Article}
}

@inproceedings{RN910,
   author = {Dong, Xiaoyi and Bao, Jianmin and Chen, Dongdong and Zhang, Weiming and Yu, Nenghai and Yuan, Lu and Chen, Dong and Guo, Baining},
   title = {Cswin transformer: A general vision transformer backbone with cross-shaped windows},
   booktitle = {Proceedings of the IEEE conference on computer vision and pattern recognition},
   pages = {12124-12134},
   year = {2022},
   type = {Conference Proceedings}
}

@article{RN745,
   author = {Dosovitskiy, Alexey},
   title = {An image is worth 16x16 words: Transformers for image recognition at scale},
   journal = {arXiv preprint arXiv:2010.11929},
   year = {2020},
   type = {Journal Article}
}

@article{RN893,
   author = {Dumoulin, Serge O},
   title = {Layers of neuroscience},
   journal = {Neuron},
   volume = {96},
   number = {6},
   pages = {1205-1206},
   ISSN = {0896-6273},
   year = {2017},
   type = {Journal Article}
}

@article{RN759,
   author = {Eagleman, David M},
   title = {Visual illusions and neurobiology},
   journal = {Nature Reviews Neuroscience},
   volume = {2},
   number = {12},
   pages = {920-926},
   ISSN = {1471-003X},
   year = {2001},
   type = {Journal Article}
}

@article{RN756,
   author = {Fan, Jinyu and Zeng, Yi},
   title = {Challenging deep learning models with image distortion based on the abutting grating illusion},
   journal = {Patterns},
   volume = {4},
   number = {3},
   ISSN = {2666-3899},
   year = {2023},
   type = {Journal Article}
}

@article{RN897,
   author = {Felleman, Daniel J and Van Essen, David C},
   title = {Distributed hierarchical processing in the primate cerebral cortex},
   journal = {Cerebral cortex (New York, N.Y. : 1991)},
   volume = {1},
   number = {1},
   pages = {1-47},
   ISSN = {1047-3211},
   year = {1991},
   type = {Journal Article}
}

@article{RN882,
   author = {Foxe, John J and Knight, Emily J and Myers, Evan J and Cao, Cody Zhewei and Molholm, Sophie and Freedman, Edward G},
   title = {The strength of feedback processing is associated with resistance to visual backward masking during Illusory Contour processing in adult humans},
   journal = {Neuroimage},
   volume = {259},
   pages = {119416},
   ISSN = {1053-8119},
   year = {2022},
   type = {Journal Article}
}

@inproceedings{RN619,
   author = {Fukushima, Kunihiko and Miyake, Sei},
   title = {Neocognitron: A self-organizing neural network model for a mechanism of visual pattern recognition},
   booktitle = {Competition and Cooperation in Neural Nets},
   pages = {267-285},
   year = {1982},
   type = {Conference Proceedings}
}

@article{RN765,
   author = {Fuss, Theodora and Bleckmann, Horst and Schluessel, Vera},
   title = {The brain creates illusions not just for us: sharks (Chiloscyllium griseum) can “see the magic” as well},
   journal = {Frontiers in Neural Circuits},
   volume = {8},
   pages = {24},
   ISSN = {1662-5110},
   year = {2014},
   type = {Journal Article}
}

@article{RN896,
   author = {Galakhova, Anna A and Hunt, Sarah and Wilbers, René and Heyer, Djai B and de Kock, Christiaan PJ and Mansvelder, Huibert D and Goriounova, Natalia A},
   title = {Evolution of cortical neurons supporting human cognition},
   journal = {Trends in cognitive sciences},
   volume = {26},
   number = {11},
   pages = {909-922},
   ISSN = {1364-6613},
   year = {2022},
   type = {Journal Article}
}

@article{RN787,
   author = {Grosof, David H and Shapley, Robert M and Hawken, Michael J},
   title = {Macaque VI neurons can signal ‘illusory’contours},
   journal = {Nature},
   volume = {365},
   number = {6446},
   pages = {550-552},
   ISSN = {0028-0836},
   year = {1993},
   type = {Journal Article}
}

@article{RN918,
   author = {Grossberg, Stephen},
   title = {How visual illusions illuminate complementary brain processes: illusory depth from brightness and apparent motion of illusory contours},
   journal = {Frontiers in human neuroscience},
   volume = {8},
   pages = {854},
   ISSN = {1662-5161},
   year = {2014},
   type = {Journal Article}
}

@inproceedings{RN608,
   author = {He, Kaiming and Zhang, Xiangyu and Ren, Shaoqing and Sun, Jian},
   title = {Deep residual learning for image recognition},
   booktitle = {Proceedings of the IEEE conference on computer vision and pattern recognition},
   pages = {770-778},
   year = {2016},
   type = {Conference Proceedings}
}

@article{RN676,
   author = {Hendrycks, Dan and Gimpel, Kevin},
   title = {Gaussian error linear units (gelus)},
   journal = {arXiv preprint arXiv:1606.08415},
   year = {2016},
   type = {Journal Article}
}

@article{RN887,
   author = {Hermann, Katherine L and Firestone, Chaz},
   title = {Shape bias at a glance: Comparing human and machine vision on equal terms},
   journal = {Journal of Vision},
   volume = {22},
   number = {14},
   pages = {3255-3255},
   ISSN = {1534-7362},
   year = {2022},
   type = {Journal Article}
}

@article{RN863,
   author = {Howard, Andrew G and Zhu, Menglong and Chen, Bo and Kalenichenko, Dmitry and Wang, Weijun and Weyand, Tobias and Andreetto, Marco and Adam, Hartwig},
   title = {Mobilenets: Efficient convolutional neural networks for mobile vision applications},
   journal = {arXiv preprint arXiv:1704.04861},
   year = {2017},
   type = {Journal Article}
}

@article{RN344,
   author = {Hubel, David H and Wiesel, Torsten N},
   title = {Receptive fields, binocular interaction and functional architecture in the cat's visual cortex},
   journal = {The Journal of Physiology},
   volume = {160},
   number = {1},
   pages = {106-154},
   ISSN = {1469-7793},
   year = {1962},
   type = {Journal Article}
}

@article{RN926,
   author = {Jesus, André and Zito, Claudio and Tortorici, Claudio and Roura, Eloy and De Masi, Giulia},
   title = {Underwater object classification and detection: first results and open challenges},
   journal = {OCEANS 2022-chennai},
   pages = {1-6},
   ISSN = {1665418214},
   year = {2022},
   type = {Journal Article}
}

@article{RN881,
   author = {Kalar, Donald J and Garrigan, Patrick and Wickens, Thomas D and Hilger, James D and Kellman, Philip J},
   title = {A unified model of illusory and occluded contour interpolation},
   journal = {Vision research},
   volume = {50},
   number = {3},
   pages = {284-299},
   ISSN = {0042-6989},
   year = {2010},
   type = {Journal Article}
}

@article{RN764,
   author = {Kanizsa, Gaetano},
   title = {Margini quasi-percettivi in campi con stimolazione omogenea},
   journal = {Rivista di psicologia},
   volume = {49},
   number = {1},
   pages = {7-30},
   year = {1955},
   type = {Journal Article}
}

@article{RN775,
   author = {Kanizsa, Gaetano},
   title = {Subjective contours},
   journal = {Scientific American},
   volume = {234},
   number = {4},
   pages = {48-53},
   ISSN = {0036-8733},
   year = {1976},
   type = {Journal Article}
}

@article{RN927,
   author = {Ke, Lei and Tai, Yu-Wing and Tang, Chi-Keung},
   title = {Occlusion-aware instance segmentation via bilayer network architectures},
   journal = {IEEE Transactions on Pattern Analysis and Machine Intelligence},
   volume = {45},
   number = {8},
   pages = {10197-10211},
   ISSN = {0162-8828},
   year = {2023},
   type = {Journal Article}
}

@inproceedings{RN922,
   author = {Kirillov, Alexander and Mintun, Eric and Ravi, Nikhila and Mao, Hanzi and Rolland, Chloe and Gustafson, Laura and Xiao, Tete and Whitehead, Spencer and Berg, Alexander C and Lo, Wan-Yen},
   title = {Segment anything},
   booktitle = {Proceedings of the IEEE international conference on computer vision},
   pages = {4015-4026},
   year = {2023},
   type = {Conference Proceedings}
}

@article{RN795,
   author = {Knebel, Jean-François and Murray, Micah M},
   title = {Towards a resolution of conflicting models of illusory contour processing in humans},
   journal = {Neuroimage},
   volume = {59},
   number = {3},
   pages = {2808-2817},
   ISSN = {1053-8119},
   year = {2012},
   type = {Journal Article}
}

@article{RN885,
   author = {Lake, Brenden M and Ullman, Tomer D and Tenenbaum, Joshua B and Gershman, Samuel J},
   title = {Building machines that learn and think like people},
   journal = {Behavioral and brain sciences},
   volume = {40},
   pages = {e253},
   ISSN = {0140-525X},
   year = {2017},
   type = {Journal Article}
}

@article{RN761,
   author = {Lesher, Gregory W},
   title = {Illusory contours: Toward a neurally based perceptual theory},
   journal = {Psychonomic Bulletin \& Review},
   volume = {2},
   pages = {279-321},
   ISSN = {1069-9384},
   year = {1995},
   type = {Journal Article}
}

@article{RN898,
   author = {Levinson, M and Podvalny, E and Baete, SH and He, BJ},
   title = {Cortical and subcortical signatures of conscious object recognition},
   journal = {Nature Communications},
   volume = {12},
   number = {1},
   pages = {1-16},
   year = {2021},
   type = {Journal Article}
}

@inproceedings{RN809,
   author = {Liu, Ze and Lin, Yutong and Cao, Yue and Hu, Han and Wei, Yixuan and Zhang, Zheng and Lin, Stephen and Guo, Baining},
   title = {Swin transformer: Hierarchical vision transformer using shifted windows},
   booktitle = {IEEE International Conference on Computer Vision},
   pages = {10012-10022},
   year = {2021},
   type = {Conference Proceedings}
}

@inproceedings{RN907,
   author = {Liu, Zhuang and Mao, Hanzi and Wu, Chao-Yuan and Feichtenhofer, Christoph and Darrell, Trevor and Xie, Saining},
   title = {A convnet for the 2020s},
   booktitle = {Proceedings of the IEEE conference on computer vision and pattern recognition},
   pages = {11976-11986},
   year = {2022},
   type = {Conference Proceedings}
}

@article{RN912,
   author = {Loshchilov, Ilya and Hutter, Frank},
   title = {Decoupled weight decay regularization},
   journal = {arXiv preprint arXiv:1711.05101},
   year = {2017},
   type = {Journal Article}
}

@inproceedings{RN805,
   author = {Lotter, William and Kreiman, Gabriel and Cox, David},
   title = {Deep predictive coding networks for video prediction and unsupervised learning},
   booktitle = {International Conference on Learning Representations},
   year = {2017},
   type = {Conference Proceedings}
}

@article{RN801,
   author = {Lotter, William and Kreiman, Gabriel and Cox, David},
   title = {A neural network trained for prediction mimics diverse features of biological neurons and perception},
   journal = {Nature machine intelligence},
   volume = {2},
   number = {4},
   pages = {210-219},
   ISSN = {2522-5839},
   year = {2020},
   type = {Journal Article}
}

@article{RN900,
   author = {Markov, Nikola T and Ercsey-Ravasz, Maria M and Ribeiro Gomes, AR and Lamy, Camille and Magrou, Loic and Vezoli, Julien and Misery, Pierre and Falchier, Arnaud and Quilodran, Rene and Gariel, Marie-Alice},
   title = {A weighted and directed interareal connectivity matrix for macaque cerebral cortex},
   journal = {Cerebral cortex},
   volume = {24},
   number = {1},
   pages = {17-36},
   ISSN = {1460-2199},
   year = {2014},
   type = {Journal Article}
}

@article{RN647,
   author = {Markov, Nikola T and Vezoli, Julien and Chameau, Pascal and Falchier, Arnaud and Quilodran, René and Huissoud, Cyril and Lamy, Camille and Misery, Pierre and Giroud, Pascale and Ullman, Shimon},
   title = {Anatomy of hierarchy: feedforward and feedback pathways in macaque visual cortex},
   journal = {Journal of Comparative Neurology},
   volume = {522},
   number = {1},
   pages = {225-259},
   ISSN = {0021-9967},
   year = {2014},
   type = {Journal Article}
}

@inproceedings{RN502,
   author = {Mottaghi, Roozbeh and Chen, Xianjie and Liu, Xiaobai and Cho, Nam-Gyu and Lee, Seong-Whan and Fidler, Sanja and Urtasun, Raquel and Yuille, Alan},
   title = {The role of context for object detection and semantic segmentation in the wild},
   booktitle = {Proceedings of the IEEE Conference on Computer Vision and Pattern Recognition},
   pages = {891-898},
   year = {2014},
   type = {Conference Proceedings}
}

@article{RN894,
   author = {Mountcastle, Vernon B},
   title = {The columnar organization of the neocortex},
   journal = {Brain: a journal of neurology},
   volume = {120},
   number = {4},
   pages = {701-722},
   ISSN = {1460-2156},
   year = {1997},
   type = {Journal Article}
}

@article{RN774,
   author = {Murray, Micah M and Herrmann, Christoph S},
   title = {Illusory contours: a window onto the neurophysiology of constructing perception},
   journal = {Trends in cognitive sciences},
   volume = {17},
   number = {9},
   pages = {471-481},
   ISSN = {1364-6613},
   year = {2013},
   type = {Journal Article}
}

@article{RN776,
   author = {Nieder, Andreas},
   title = {Seeing more than meets the eye: processing of illusory contours in animals},
   journal = {Journal of Comparative Physiology A},
   volume = {188},
   pages = {249-260},
   ISSN = {0340-7594},
   year = {2002},
   type = {Journal Article}
}

@article{RN766,
   author = {Nieder, Andreas and Wagner, Hermann},
   title = {Perception and neuronal coding of subjective contours in the owl},
   journal = {Nature neuroscience},
   volume = {2},
   number = {7},
   pages = {660-663},
   ISSN = {1546-1726},
   year = {1999},
   type = {Journal Article}
}

@article{RN923,
   author = {Otsu, N.},
   title = {A Threshold Selection Method from Gray-Level Histograms},
   journal = {IEEE Transactions on Systems, Man, and Cybernetics},
   volume = {9},
   number = {1},
   pages = {62-66},
   ISSN = {2168-2909},
   year = {1979},
   type = {Journal Article}
}

@article{RN792,
   author = {Pak, Alexandr and Ryu, Esther and Li, Claudia and Chubykin, Alexander A},
   title = {Top-down feedback controls the cortical representation of illusory contours in mouse primary visual cortex},
   journal = {Journal of Neuroscience},
   volume = {40},
   number = {3},
   pages = {648-660},
   ISSN = {0270-6474},
   year = {2020},
   type = {Journal Article}
}

@article{RN767,
   author = {Pan, Yanxia and Chen, Minggui and Yin, Jiapeng and An, Xu and Zhang, Xian and Lu, Yiliang and Gong, Hongliang and Li, Wu and Wang, Wei},
   title = {Equivalent representation of real and illusory contours in macaque V4},
   journal = {Journal of Neuroscience},
   volume = {32},
   number = {20},
   pages = {6760-6770},
   ISSN = {0270-6474},
   year = {2012},
   type = {Journal Article}
}

@article{RN803,
   author = {Pang, Zhaoyang and O’May, Callum Biggs and Choksi, Bhavin and VanRullen, Rufin},
   title = {Predictive coding feedback results in perceived illusory contours in a recurrent neural network},
   journal = {Neural Networks},
   volume = {144},
   pages = {164-175},
   ISSN = {0893-6080},
   year = {2021},
   type = {Journal Article}
}

@article{RN888,
   author = {Peterhans, Esther and von der Heydt, Rüdiger},
   title = {Mechanisms of contour perception in monkey visual cortex. II. Contours bridging gaps},
   journal = {Journal of Neuroscience},
   volume = {9},
   number = {5},
   pages = {1749-1763},
   ISSN = {0270-6474},
   year = {1989},
   type = {Journal Article}
}

@inproceedings{RN734,
   author = {Pu, Mengyang and Huang, Yaping and Liu, Yuming and Guan, Qingji and Ling, Haibin},
   title = {Edter: Edge detection with transformer},
   booktitle = {Proceedings of the IEEE conference on computer vision and pattern recognition},
   pages = {1402-1412},
   year = {2022},
   type = {Conference Proceedings}
}

@article{RN806,
   author = {Rao, Rajesh PN and Ballard, Dana H},
   title = {Predictive coding in the visual cortex: a functional interpretation of some extra-classical receptive-field effects},
   journal = {Nature neuroscience},
   volume = {2},
   number = {1},
   pages = {79-87},
   ISSN = {1546-1726},
   year = {1999},
   type = {Journal Article}
}

@article{RN797,
   author = {Ringach, Dario L. and Shapley, Robert},
   title = {Spatial and Temporal Properties of Illusory Contours and Amodal Boundary Completion},
   journal = {Vision Research},
   volume = {36},
   number = {19},
   pages = {3037-3050},
   ISSN = {0042-6989},
   year = {1996},
   type = {Journal Article}
}

@article{RN811,
   author = {Roelfsema, Pieter R},
   title = {Solving the binding problem: Assemblies form when neurons enhance their firing rate—they don’t need to oscillate or synchronize},
   journal = {Neuron},
   volume = {111},
   number = {7},
   pages = {1003-1019},
   ISSN = {0896-6273},
   year = {2023},
   type = {Journal Article}
}

@article{RN744,
   author = {Russakovsky, Olga and Deng, Jia and Su, Hao and Krause, Jonathan and Satheesh, Sanjeev and Ma, Sean and Huang, Zhiheng and Karpathy, Andrej and Khosla, Aditya and Bernstein, Michael},
   title = {Imagenet large scale visual recognition challenge},
   journal = {International journal of computer vision},
   volume = {115},
   pages = {211-252},
   ISSN = {0920-5691},
   year = {2015},
   type = {Journal Article}
}

@article{RN791,
   author = {Sáry, Gy and Chadaide, Zoltán and Tompa, Tamás and Köteles, K and Kovács, GY and Benedek, G},
   title = {Illusory shape representation in the monkey inferior temporal cortex},
   journal = {European Journal of Neuroscience},
   volume = {25},
   number = {8},
   pages = {2558-2564},
   ISSN = {0953-816X},
   year = {2007},
   type = {Journal Article}
}

@article{RN762,
   author = {Sary, Gy and Köteles, K and Kaposvári, Péter and Lenti, L and Csifcsák, Gábor and Frankó, E and Benedek, György and Tompa, Tamás},
   title = {The representation of Kanizsa illusory contours in the monkey inferior temporal cortex},
   journal = {European Journal of Neuroscience},
   volume = {28},
   number = {10},
   pages = {2137-2146},
   ISSN = {0953-816X},
   year = {2008},
   type = {Journal Article}
}

@article{RN823,
   author = {Schulze Buschoff, Luca M and Akata, Elif and Bethge, Matthias and Schulz, Eric},
   title = {Visual cognition in multimodal large language models},
   journal = {Nature Machine Intelligence},
   volume = {7},
   number = {1},
   pages = {96-106},
   ISSN = {2522-5839},
   year = {2025},
   type = {Journal Article}
}

@article{RN760,
   author = {Schumann, Friedrich},
   title = {Einige Beobachtungen über die Zusammenfassung von Gesichtseindrücken zu Einheiten},
   journal = {Zeitschrift Für Psychologie und Physiologie der Sinnesorgane},
   volume = {23},
   pages = {1-32},
   year = {1900},
   type = {Journal Article}
}

@inproceedings{RN917,
   author = {Selvaraju, Ramprasaath R and Cogswell, Michael and Das, Abhishek and Vedantam, Ramakrishna and Parikh, Devi and Batra, Dhruv},
   title = {Grad-cam: Visual explanations from deep networks via gradient-based localization},
   booktitle = {Proceedings of the IEEE international conference on computer vision},
   pages = {618-626},
   year = {2017},
   type = {Conference Proceedings}
}

@article{RN789,
   author = {Sheth, Bhavin R and Sharma, Jitendra and Rao, S Chenchal and Sur, Mriganka},
   title = {Orientation maps of subjective contours in visual cortex},
   journal = {Science},
   volume = {274},
   number = {5295},
   pages = {2110-2115},
   ISSN = {0036-8075},
   year = {1996},
   type = {Journal Article}
}

@article{RN884,
   author = {Shishikura, Motofumi and Machida, Itsuki and Tamura, Hiroshi and Sakai, Ko},
   title = {Local contour features contribute to figure-ground segregation in monkey V4 neural populations and human perception},
   journal = {Neural Networks},
   volume = {181},
   pages = {106821},
   ISSN = {0893-6080},
   year = {2025},
   type = {Journal Article}
}

@article{RN901,
   author = {Shushruth, S and Ichida, Jennifer M and Levitt, Jonathan B and Angelucci, Alessandra},
   title = {Comparison of spatial summation properties of neurons in macaque V1 and V2},
   journal = {Journal of neurophysiology},
   volume = {102},
   number = {4},
   pages = {2069-2083},
   ISSN = {0022-3077},
   year = {2009},
   type = {Journal Article}
}

@inproceedings{RN490,
   author = {Simonyan, Karen and Zisserman, Andrew},
   title = {Very deep convolutional networks for large-scale image recognition},
   booktitle = {International Conference on Learning Representations},
   pages = {1-14},
   year = {2015},
   type = {Conference Proceedings}
}

@article{RN880,
   author = {Smith, Gabriella E and Chouinard, Philippe A and Byosiere, Sarah-Elizabeth},
   title = {If I fits I sits: A citizen science investigation into illusory contour susceptibility in domestic cats (Felis silvestris catus)},
   journal = {Applied Animal Behaviour Science},
   volume = {240},
   pages = {105338},
   ISSN = {0168-1591},
   year = {2021},
   type = {Journal Article}
}

@article{RN924,
   author = {Soriano, Manuel and Spillmann, Lothar and Bach, Michael},
   title = {The abutting grating illusion},
   journal = {Vision research},
   volume = {36},
   number = {1},
   pages = {109-116},
   ISSN = {0042-6989},
   year = {1996},
   type = {Journal Article}
}

@article{RN748,
   author = {Suzuki, Mototaka and Pennartz, Cyriel MA and Aru, Jaan},
   title = {How deep is the brain? The shallow brain hypothesis},
   journal = {Nature Reviews Neuroscience},
   volume = {24},
   number = {12},
   pages = {778-791},
   ISSN = {1471-003X},
   year = {2023},
   type = {Journal Article}
}

@article{RN786,
   author = {von der Heydt, Rüdiger and Peterhans, Esther},
   title = {Mechanisms of contour perception in monkey visual cortex. I. Lines of pattern discontinuity},
   journal = {Journal of Neuroscience},
   volume = {9},
   number = {5},
   pages = {1731-1748},
   ISSN = {0270-6474},
   year = {1989},
   type = {Journal Article}
}

@article{RN783,
   author = {von der Heydt, Rüdiger and Peterhans, Esther and Baumgartner, Gunter},
   title = {Illusory contours and cortical neuron responses},
   journal = {Science},
   volume = {224},
   number = {4654},
   pages = {1260-1262},
   ISSN = {0036-8075},
   year = {1984},
   type = {Journal Article}
}

@article{RN899,
   author = {Wagstyl, Konrad and Larocque, Stéphanie and Cucurull, Guillem and Lepage, Claude and Cohen, Joseph Paul and Bludau, Sebastian and Palomero-Gallagher, Nicola and Lewis, Lindsay B and Funck, Thomas and Spitzer, Hannah},
   title = {BigBrain 3D atlas of cortical layers: Cortical and laminar thickness gradients diverge in sensory and motor cortices},
   journal = {PLoS biology},
   volume = {18},
   number = {4},
   pages = {e3000678},
   ISSN = {1544-9173},
   year = {2020},
   type = {Journal Article}
}

@article{RN902,
   author = {Wandell, Brian A and Winawer, Jonathan},
   title = {Computational neuroimaging and population receptive fields},
   journal = {Trends in cognitive sciences},
   volume = {19},
   number = {6},
   pages = {349-357},
   ISSN = {1364-6613},
   year = {2015},
   type = {Journal Article}
}

@article{RN906,
   author = {Wang, Wenbo and Zhou, Tiangang and Chen, Lin and Huang, Yan},
   title = {A subcortical magnocellular pathway is responsible for the fast processing of topological properties of objects: A transcranial magnetic stimulation study},
   journal = {Human Brain Mapping},
   volume = {44},
   number = {4},
   pages = {1617-1628},
   ISSN = {1065-9471},
   year = {2023},
   type = {Journal Article}
}

@inproceedings{RN675,
   author = {Woo, Sanghyun and Park, Jongchan and Lee, Joon-Young and Kweon, In So},
   title = {CBAM: Convolutional Block Attention Module},
   booktitle = {European Conference on Computer Vision},
   series = {Computer Vision – ECCV 2018},
   pages = {3-19},
   ISBN = {978-3-030-01234-2},
   year = {2018},
   type = {Conference Proceedings}
}

@article{RN763,
   author = {Wyzisk, Katja and Neumeyer, Christa},
   title = {Perception of illusory surfaces and contours in goldfish},
   journal = {Visual neuroscience},
   volume = {24},
   number = {3},
   pages = {291-298},
   ISSN = {1469-8714},
   year = {2007},
   type = {Journal Article}
}

@inproceedings{RN908,
   author = {Xia, Chunlong and Wang, Xinliang and Lv, Feng and Hao, Xin and Shi, Yifeng},
   title = {Vit-comer: Vision transformer with convolutional multi-scale feature interaction for dense predictions},
   booktitle = {Proceedings of the IEEE conference on computer vision and pattern recognition},
   pages = {5493-5502},
   year = {2024},
   type = {Conference Proceedings}
}

@article{RN920,
   author = {Xiao, Han and Rasul, Kashif and Vollgraf, Roland},
   title = {Fashion-mnist: a novel image dataset for benchmarking machine learning algorithms},
   journal = {arXiv preprint arXiv:1708.07747},
   year = {2017},
   type = {Journal Article}
}

@inproceedings{RN481,
   author = {Xie, Saining and Tu, Zhuowen},
   title = {Holistically-nested edge detection},
   booktitle = {International Conference on Computer Vision},
   pages = {1395-1403},
   year = {2015},
   type = {Conference Proceedings}
}

@inproceedings{RN904,
   author = {Yu, Fisher Yu and Koltun, Vladlen},
   title = {Multi-Scale Context Aggregation by Dilated Convolutions},
   booktitle = {International Conference on Learning Representations},
   pages = {1-13},
   year = {2015},
   type = {Conference Proceedings}
}

@inproceedings{RN810,
   author = {Yu, Weihao and Wang, Xinchao},
   title = {MambaOut: Do We Really Need Mamba for Vision?},
   booktitle = {Proceedings of the IEEE conference on computer vision and pattern recognition},
   year = {2025},
   type = {Conference Proceedings}
}

@article{RN747,
   author = {Zhang, Xiao and Lin, Chuan and Li, Fuzhang and Cao, Yijun and Li, Yongjie},
   title = {LVP-net: A deep network of learning visual pathway for edge detection},
   journal = {Image and Vision Computing},
   volume = {147},
   pages = {105078},
   ISSN = {0262-8856},
   year = {2024},
   type = {Journal Article}
}

@inproceedings{RN909,
   author = {Zhang, Xiangyu and Zhou, Xinyu and Lin, Mengxiao and Sun, Jian},
   title = {Shufflenet: An extremely efficient convolutional neural network for mobile devices},
   booktitle = {Proceedings of the IEEE conference on computer vision and pattern recognition},
   pages = {6848-6856},
   year = {2018},
   type = {Conference Proceedings}
}

@inproceedings{RN804,
   author = {Zhang, Yichi and Pan, Jiayi and Zhou, Yuchen and Pan, Rui and Chai, Joyce},
   title = {Grounding Visual Illusions in Language: Do Vision-Language Models Perceive Illusions Like Humans?},
   booktitle = {Conference on Empirical Methods in Natural Language Processing},
   pages = {5718-5728},
   type = {Conference Proceedings},
   year= {2023}
}

@inproceedings{RN746,
   author = {Zhou, Caixia and Huang, Yaping and Pu, Mengyang and Guan, Qingji and Deng, Ruoxi and Ling, Haibin},
   title = {MuGE: Multiple Granularity Edge Detection},
   booktitle = {Proceedings of the IEEE conference on computer vision and pattern recognition},
   pages = {25952-25962},
   year = {2024},
   type = {Conference Proceedings}
}

@article{RN773,
   author = {Cheng, Fan L and Horikawa, Tomoyasu and Majima, Kei and Tanaka, Misato and Abdelhack, Mohamed and Aoki, Shuntaro C and Hirano, Jin and Kamitani, Yukiyasu},
   title = {Reconstructing visual illusory experiences from human brain activity},
   journal = {Science Advances},
   volume = {9},
   number = {46},
   pages = {eadj3906},
   ISSN = {2375-2548},
   year = {2023},
   type = {Journal Article}
}

@article{RN929,
   author = {Lee, Tai Sing and Nguyen, My},
   title = {Dynamics of subjective contour formation in the early visual cortex},
   journal = {Proceedings of the National Academy of Sciences},
   volume = {98},
   number = {4},
   pages = {1907-1911},
   ISSN = {0027-8424},
   year = {2001},
   type = {Journal Article}
}

@article{RN930,
   author = {Lee, Tai Sing},
   title = {The nature of illusory contour computation},
   journal = {Neuron},
   volume = {33},
   number = {5},
   pages = {667-668},
   ISSN = {0896-6273},
   year = {2002},
   type = {Journal Article}
}

@article{RN931,
   author = {Huxlin, Krystel R and Saunders, Richard C and Marchionini, Deanna and Pham, Hong-An and Merigan, William H},
   title = {Perceptual deficits after lesions of inferotemporal cortex in macaques},
   journal = {Cerebral Cortex},
   volume = {10},
   number = {7},
   pages = {671-683},
   ISSN = {1460-2199},
   year = {2000},
   type = {Journal Article}
}

@article{RN790,
   author = {De Weerd, Peter and Desimone, Robert and Ungerleider, Leslie G},
   title = {Cue-dependent deficits in grating orientation discrimination after V4 lesions in macaques},
   journal = {Visual neuroscience},
   volume = {13},
   number = {3},
   pages = {529-538},
   ISSN = {1469-8714},
   year = {1996},
   type = {Journal Article}
}

@article{RN796,
   author = {Mendola, Janine D and Dale, Anders M and Fischl, Bruce and Liu, Arthur K and Tootell, Roger BH},
   title = {The representation of illusory and real contours in human cortical visual areas revealed by functional magnetic resonance imaging},
   journal = {Journal of Neuroscience},
   volume = {19},
   number = {19},
   pages = {8560-8572},
   ISSN = {0270-6474},
   year = {1999},
   type = {Journal Article}
}

@article{RN793,
   author = {Murray, Micah M and Wylie, Glenn R and Higgins, Beth A and Javitt, Daniel C and Schroeder, Charles E and Foxe, John J},
   title = {The spatiotemporal dynamics of illusory contour processing: combined high-density electrical mapping, source analysis, and functional magnetic resonance imaging},
   journal = {Journal of Neuroscience},
   volume = {22},
   number = {12},
   pages = {5055-5073},
   ISSN = {0270-6474},
   year = {2002},
   type = {Journal Article}
}

@article{RN935,
   author = {Bergmann, J and Petro, LS and Abbatecola, C and Li, MS and Morgan, AT and Muckli, L},
   title = {Cortical depth profiles in primary visual cortex for illusory and imaginary experiences},
   journal = {Nature Communications},
   volume = {15},
   number = {1},
   pages = {1-13},
   year = {2024},
   type = {Journal Article}
}

@article{RN932,
   author = {Frégnac, Yves and Bathellier, Brice},
   title = {Cortical correlates of low-level perception: from neural circuits to percepts},
   journal = {Neuron},
   volume = {88},
   number = {1},
   pages = {110-126},
   ISSN = {0896-6273},
   year = {2015},
   type = {Journal Article}
}

@inproceedings{RN936,
   author = {Geirhos, Robert and Rubisch, Patricia and Michaelis, Claudio and Bethge, Matthias and Wichmann, Felix A and Brendel, Wieland},
   title = {ImageNet-trained CNNs are biased towards texture; increasing shape bias improves accuracy and robustness},
   booktitle = {International conference on learning representations},
   year = {2019},
   type = {Conference Proceedings}
}

@article{RN933,
   author = {Lu, Yiliang and Yin, Jiapeng and Chen, Zheyuan and Gong, Hongliang and Liu, Ye and Qian, Liling and Li, Xiaohong and Liu, Rui and Andolina, Ian Max and Wang, Wei},
   title = {Revealing detail along the visual hierarchy: neural clustering preserves acuity from V1 to V4},
   journal = {Neuron},
   volume = {98},
   number = {2},
   pages = {417-428. e3},
   ISSN = {0896-6273},
   year = {2018},
   type = {Journal Article}
}

@article{RN934,
   author = {Garcia-Marin, Virginia and Kelly, Jenna G and Hawken, Michael J},
   title = {Neuronal composition of processing modules in human V1: laminar density for neuronal and non-neuronal populations and a comparison with macaque},
   journal = {Cerebral Cortex},
   volume = {34},
   number = {2},
   pages = {bhad512},
   ISSN = {1047-3211},
   year = {2024},
   type = {Journal Article}
}
}

\end{document}